\documentclass[10pt,journal,compsoc]{IEEEtran}
\usepackage[utf8]{inputenc} 
\usepackage[T1]{fontenc}  

\usepackage[breaklinks=true,
            colorlinks,
            linkcolor = red,
            urlcolor  = purple, 
            citecolor = blue,
            bookmarks = black]{hyperref}  
\usepackage{url}  
\usepackage{amssymb}   
\usepackage{booktabs} 
\usepackage{amsfonts}  
\usepackage{nicefrac}  
\usepackage{microtype}  
\usepackage{wrapfig}
\usepackage{bbding}
\usepackage{appendix}
\usepackage{graphicx}
\usepackage{amsmath}
\usepackage{amsthm}
\usepackage{subcaption}
\usepackage{float}
\usepackage{appendix}
\usepackage{xcolor}
\usepackage{multirow}

\usepackage{bm}
\usepackage{bbm}
\usepackage{makecell}
\usepackage{mathtools}
\usepackage{algorithmic}
\usepackage{algorithm}
\usepackage[algo2e,linesnumbered]{algorithm2e}

\newcommand{\etal}{\textit{et al.}\ }
\newcommand{\ie}{\textit{i.e.}\ }
\newcommand{\eg}{\textit{e}.\textit{g}.}

\SetArgSty{textrm}
\ifCLASSOPTIONcompsoc
  \usepackage[nocompress]{cite}
\else
  \usepackage{cite}
\fi

\ifCLASSINFOpdf
\else
\fi

\begin{document}

%
\title{Tackling Noisy Labels with Network Parameter Additive Decomposition}
%
%
%
%

\author{Jingyi Wang, Xiaobo Xia, Long Lan$^\dagger$, \IEEEmembership{Member, IEEE}, Xinghao Wu, Jun Yu, Wenjing Yang, \\Bo Han, and Tongliang Liu, \IEEEmembership{Senior Member, IEEE}
\thanks{$^\dagger$\quad \ Corresponding author.}
\IEEEcompsocitemizethanks{
\IEEEcompsocthanksitem Jingyi Wang, Long Lan, and Wenjing Yang are with the Department of Intelligent Data Science, College of Computer Science and Technology, National University of Defense Technology, Changsha, 410073, China (e-mail: wangjingyi@nudt.edu.cn;long.lan@nudt.edu.cn; wenjing.yang@nudt.edu.cn).

\IEEEcompsocthanksitem Xiaobo Xia and Tongliang Liu are with the Trustworthy Machine Learning Lab, School of Computer Science, Faculty of Engineering, University of Sydney, Darlington, NSW2008, Australia (e-mail: xiaoboxia.uni@gmail.com; tongliang.liu@sydney.edu.au).

\IEEEcompsocthanksitem Xinghao Wu is with the State Key Laboratory of Virtual Reality Technology and Systems, School of Computer Science and Engineering, Beihang University, Beijing, 100191, China (e-mail: wuxinghao@buaa.edu.cn).

\IEEEcompsocthanksitem{Jun Yu is with the Department of Automation, University of Science and Technology of China, Hefei, 230026, China (e-mail: harryjun@ustc.edu.cn).}
\IEEEcompsocthanksitem Bo Han is with the Department of Computer Science, Hong Kong Baptist University, Hong Kong, China (e-mail: bhanml@comp.hkbu.edu.hk).
}

}

\markboth{Tackling Noisy Labels with Network Parameter Additive Decomposition}%
{Tackling Noisy Labels with Network Parameter Additive Decomposition}

\IEEEtitleabstractindextext{%
\begin{abstract}
 Given data with noisy labels, over-parameterized deep networks suffer overfitting mislabeled data, resulting in poor generalization. The memorization effect of deep networks shows that although the networks have the ability to memorize all noisy data, they would first memorize clean training data, and then gradually memorize mislabeled training data. A simple and effective method that exploits the memorization effect to combat noisy labels is early stopping. 
 However, early stopping cannot distinguish the memorization of clean data and mislabeled data, resulting in the network still inevitably overfitting mislabeled data in the early training stage.
 In this paper, to decouple the memorization of clean data and mislabeled data, and further reduce the side effect of mislabeled data, we perform additive decomposition on network parameters. Namely, all parameters are additively decomposed into two groups, \ie, parameters $\mathbf{w}$ are decomposed as $\mathbf{w}=\bm{\sigma}+\bm{\gamma}$. Afterward, the parameters $\bm{\sigma}$ are considered to memorize clean data, while the parameters $\bm{\gamma}$ are considered to memorize mislabeled data. Benefiting from the memorization effect, the updates of the parameters $\bm{\sigma}$ are encouraged to fully memorize clean data in early training, and then discouraged with the increase of training epochs to reduce interference of mislabeled data. The updates of the parameters $\bm{\gamma}$ are the opposite. In testing, only the parameters $\bm{\sigma}$ are employed to enhance generalization. Extensive experiments on both simulated and real-world benchmarks confirm the superior performance of our method. The implementation is available at: \href{https://github.com/wangjingyi9924/TNLPAD}{https://github.com/wangjingyi9924/TNLPAD}.
\end{abstract}
}

\maketitle

\IEEEdisplaynontitleabstractindextext

\IEEEpeerreviewmaketitle

\section{Introduction}\label{sec:introduction}

\IEEEPARstart{D}{eep} networks have achieved remarkable performance in multiple tasks with high-quality human annotations~\cite{lecun2015deep,wang2019co,seo2019combinatorial,ye2021collaborative,fu2022large,yan2023adaptive,xia2023moderate,zhang2023ideal}. However, data in real life is noisy, which is always caused by the mistakes of manual and automatic annotators~\cite{zheng2020error,li2020dividemix,wu2020class2simi,yao2020searching,ding2019decode,arazo2019unsupervised,xia2023holistic}. The mistakes result in the data with noisy labels. As deep networks have large learning capacities and
strong memorization power, they will ultimately overfit mislabeled data, leading to poor generalization performance~\cite{zhang2017understanding,feng2021can,guo2018curriculumnet,song2019selfie,algan2022metalabelnet,wu2021lr}. General regularization techniques such as
dropout and weight decay cannot handle the issue well~\cite{chen2021noise,han2020sigua,li2022estimating}.

Recently, the memorization effect of deep networks~\cite{arpit2017closer, liu2020early} creates a powerful paradigm for handling noisy labels. That is, although the over-parameterized deep networks have the ability to memorize all noisy data, the memorization effect suggests that they would first memorize clean training data, which helps generalization, and then gradually memorize mislabeled data. One effective way to exploit the memorization effect for tackling noisy labels is early stopping~\cite{li2019gradient, hu2020simple}, whose idea is to stop training when the performance of the network on the validation set is (almost) no longer increasing, thus avoiding overfitting in the later iterations \cite{song2019does}. Although the idea of early stopping is simple, as shown in precedent~\cite{li2019gradient} and our experiments in Section~\ref{sec:experiments}, it is a very valid and robust method that outperforms lots of state-of-the-art (SOTA) methods in a series of label noise scenarios. 

\begin{figure}[!t]
	\centering
        \includegraphics[width=0.49\textwidth]{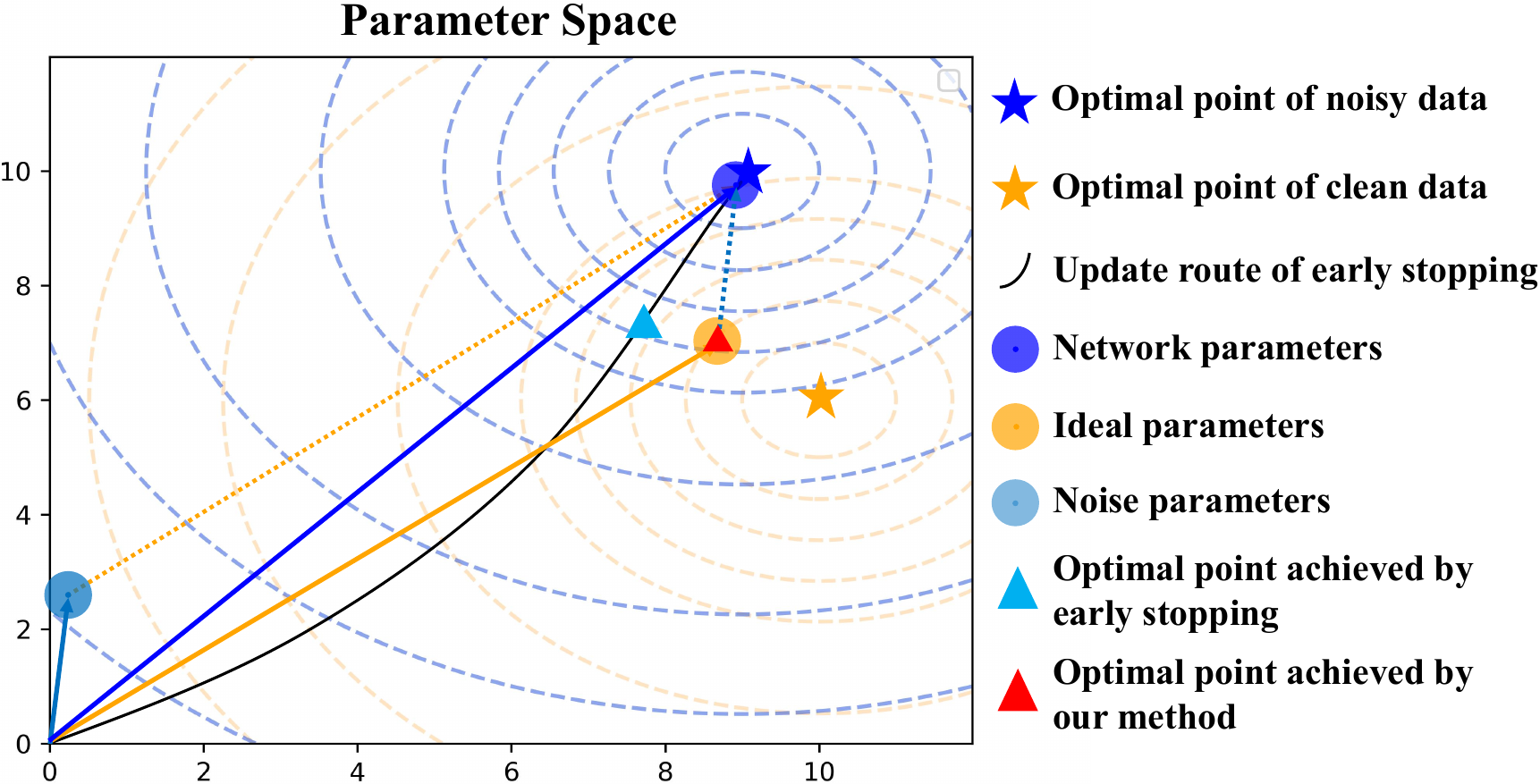}

	\caption{A toy example to illustrate the superiority of our method over the early stopping from the perspective of parameter space. The parameters of the origin network (shown as a blue circle) are decomposed into the sum of the ideal parameters (shown as a yellow circle) and the noise parameters (shown as a light blue circle). By making noise parameters absorb the partial effect of mislabeled data, and with the principle of parameter decomposition, this method can make the best ideal parameters (shown as a red triangle) closer to the optimal point of clean data, compared with the parameters trained by vanilla early stopping (shown as a blue triangle).}
	\label{fig:toyexample}
\end{figure}

However, early stopping just leverages the \textit{general tendency} of the memorization effect. As shown in Fig.~\ref{fig:toyexample}, in the early training stage, the deep network mainly memorizes clean data. It updates towards the optimal point of clean data (shown as the yellow star). As the training goes on, the network gradually memorizes mislabeled data. Its update direction is towards the optimal point of noisy data (shown as the blue star). In essence, the early stopping method simply selects the network that is closest to the optimal point of clean data in the update process (shown as the blue triangle). This method lacks control over the network update process and is easily affected by the location of the optimal points of clean and noisy data as well as the network update trajectory. The key factor affecting early stopping performance is the inability to control the network's memorization of clean and mislabeled data. In other words, in the early training stage, the network can still memorize mislabeled data, so its update deviates from the optimal point of clean data too early, resulting in poor performance. We hence raise a question: \textit{is it possible to control the training process of the network to reduce the network's memorization of mislabeled data, and make it closer to the optimal point of clean data?}

With this question in mind, we find that some related studies prove that the deep network parameters can be decomposed into several parts and each part of the parameters can focus on different aspects of information in the task \cite{yoon2019scalable}. Enlightened by this, we can decompose the network parameters into two parts in the label noise scenarios. One part is termed the ideal parameters which are used to memorize clean data, while the other part, is called noise parameters used to memorize mislabeled data. In this way, we can decouple the memorization of clean data and mislabeled data at the parameter level. The ideal parameters are less influenced by mislabeled data. The network with ideal parameters is therefore closer to the optimal point of clean data. However, since we cannot directly separate clean data and mislabeled data from datasets, how to control the two-part parameters to implement our idea is a challenge. Note that the memorization effect in learning with noisy labels shows that the deep networks would first memorize clean data and gradually mislabeled data. Inspired by this, we propose to constrain the updates of noise parameters in the early training stage and make ideal parameters fully memorize clean data. As the training goes on, the constraints on noise parameters are gradually reduced and the constraints on ideal parameters are enhanced, so that mislabeled data are mainly memorized by noise parameters. We take Fig.~\ref{fig:toyexample} to illustrate the superiority of our method.
 
Based on the above analyses, in this paper, we propose a new method that first decomposes all parameters into two parts. Then, combined with the memorization effect, we propose a time-varying loss function to dynamically control the updates of the two-part parameters. The noise parameters can then absorb part of the effect of mislabeled data. In the inference phase, we discard the noise parameters and only use the ideal parameters to obtain better network performance. The contributions of this paper are in three folds:

\begin{itemize}
    \item We propose that network parameters can be grouped into two parts with additive parameter decomposition to handle noisy labels: the ideal parameters for the memorization of clean data and the noise parameters for the memorization of mislabeled data. 
    \item Benefiting from the memorization effect of deep networks, we impose dynamic regularization constraints on the two types of parameters, which make ideal parameters fit clean data and noise parameters absorb the side effect of mislabeled data.
    \item We conduct experiments on both simulated and real-world noisy datasets, which clearly demonstrate our method achieves better performance compared with multiple SOTA methods. Comprehensive ablation studies and discussions are also provided. 
\end{itemize}

The rest of this paper is organized as follows. In Section~\ref{sec:related_work}, we review the literature related to this paper. In Section~\ref{sec:method}, we discuss the technical details of the proposed method step by step. Extensive experiments and ablation studies are presented in Section~\ref{sec:experiments}. Finally, we conclude this paper in Section~\ref{sec:conclusion}.

\section{Related Work}\label{sec:related_work}
Learning with noisy labels has gained much interest~\cite{han2020survey,yao2018deep}. Lots of works have proposed multiple methods to handle noisy labels, which can be generally categorized into two groups: methods with the memorization effect and methods without the memorization effect. In this section, we briefly review the recent related works.

\subsection{Methods with the Memorization Effect}
\textbf{Sample Selection.} In recent years, the memorization effect of deep networks proposed by Arpit \etal \cite{arpit2017closer} has been widely used in learning with noisy labels. It represents the behavior exhibited by deep networks trained on noisy datasets. Based on the memorization effect, many researchers adopt a sample selection strategy, which is called the small-loss trick \cite{jiang2018mentornet, ren2018learning, han2018co, yu2019does, wei2020combating,xia2023combating}. The core idea is to select a part of small-loss examples from the noisy dataset as clean data for updating. \cite{jiang2018mentornet} builds on curriculum learning and proposes a neural network called MentorNet, which aims to guide the training of the student network. Based on the small-loss trick, the mentor network provides a curriculum for the student network during training to focus on examples whose labels are likely to be correct. 
CoTeaching \cite{han2018co}, CoTeaching+ \cite{yu2019does}, and JoCor \cite{wei2020combating} all train two networks at the same time. CoTeaching chooses some small-loss data for each network and transmits it to the peer-to-peer network.  CoTeaching+ bridges the updated disagreement strategy to the original CoTeaching. Jocor updates two networks through co-regularization. \cite{xia2023combating} proposes a method called CoDis for training deep networks by selecting data with high-discrepancy prediction probabilities between two networks, thus maintaining network divergence and improving sample efficiency. 

\noindent\textbf{Label correction.} The methods in this group attempt to correct incorrect labels to clean labels for improving network robustness. Specifically, \cite{vahdat2017toward} proposes an undirected graphical model to represent the relationship between noisy and clean labels, and designs an objective function for training the deep structured model.
\cite{li2017learning} designs a new framework to correct noisy labels by leveraging the knowledge learned from a small clean dataset and semantic knowledge graph.
For learning deep network parameters and estimating true labels, \cite{tanaka2018joint} proposes a joint optimization framework, which can correct the labels during training by alternately updating the network parameters and labels.
\cite{yi2019probabilistic} adopts label probability distributions to supervise network learning and updates these distributions via backpropagation.
\cite{chen2020beyond} provides label correction by averaging the output of the deep network on each example throughout the training process, and then retraining the classifier using the averaged soft labels.

\noindent\textbf{Learning with parameter constraints.}
\cite{xia2021robust} makes an attempt by classifying all parameters into two classes based on the importance of the memorization of clean data, \ie, important parameters and redundant parameters. The updates of redundant parameters are limited to reduce the overfitting of mislabeled data. In this way, a parameter must be recognized to be important or redundant in binary, which will cause useful information to be lost, resulting in limited robustness improvement. 
In this paper, we propose a new idea from the perspective of additive parameter decomposition and we impose parameter constraints to strengthen the utilization of the learning tendency of the memorization effect, thus further weakening the side effect of mislabeled data and achieving better robustness.

\subsection{Methods without the Memorization Effect}
\noindent\textbf{Loss correction.} The methods in this group aim to improve the network performance by modifying the training loss~\cite{yong2023holistic}. \cite{liu2016classification} proposes to use the reweighting loss for classification with noisy labels. \cite{goldberger2016training} adds a special noise adaptation layer that connects clean labels to noisy ones to correct network outputs. In addition, \cite{patrini2017making, xia2019anchor,yao2020dual,xia2022extended,zhu2021second,shu2020meta,bucarelli2023leveraging,liu2023identifiability} estimate the noise transition matrix by revealing the transition relationship from clean labels to noisy labels and then correct the loss. In more detail, 
\cite{patrini2017making} proposes two procedures for loss correction, including the forward correction using the noise transition matrix and the back correction using the inversed transition matrix.
\cite{xia2019anchor} initializes the matrix with data points similar to anchor points, and then introduces slack variables to modify the noise transition matrix. 
In \cite{yao2020dual}, an intermediate class is designed so that the original transition matrix can be decomposed into the product of two transition matrices that are easy to estimate, thus avoiding direct estimation of the noise class posterior probability.
\cite{xia2022extended} designs an extended $T$-estimator based on the traditional transition matrix to estimate the cluster-dependent extended transition matrix by only exploiting the noisy data.
\cite{zhu2021second} proposes a second-order strategy that utilizes the estimation of several covariance terms defined between the instance-dependent noise rates and the Bayesian-optimal label to handle instance-dependent noise. 
\cite{bucarelli2023leveraging} leverages inter-annotator agreement (IAA) statistics to estimate the label noise distribution and modifies learning algorithms to be robust to the resulting noise in the labels, providing generalization bounds based on known quantities.\cite{liu2023identifiability} conducts a series of studies on the identifiability of instance-level label noise transition matrix.
Obviously, the estimation quality of the noise transition matrix plays a decisive role in tackling noisy labels. However, it is difficult to accurately estimate the matrix in the current research, especially in large classes or heavy noise~\cite{han2018co}.

\noindent\textbf{Robust loss functions.} The classical cross-entropy (CE) loss is widely used in classification problems due to its good generalization capability and fast convergence. However, CE is not robust when learning with noisy labels. Using only CE loss will overfit noisy labels and result in the degradation of the generalization performance of deep networks. Mean absolute error (MAE) loss~\cite{ghosh2017robust} has been proven to be robust to noisy labels. Combined with the advantages of the noise robustness of MAE and the implicit weighting scheme of CE, generalized cross-entropy (GCE) loss~\cite{zhang2018generalized} is proposed, which is a more general noise robust loss. Inspired by the symmetric KL-divergence, symmetric cross-entropy (SCE) loss~\cite{wang2019symmetric} is proposed to address underlearning and overfitting problems by boosting CE symmetrically with the noise-robust reverse cross-entropy (RCE). Active passive loss (APL) \cite{ma2020normalized} also combines two robust loss functions. Among them, active loss explicitly maximizes the probability of being in the labeled class, and passive loss explicitly minimizes the probability of being in other classes. Curriculum learning (CL)~\cite{lyu2019curriculum} is a surrogate loss of the 0-1 loss. CL has a tighter upper bound, and could adaptively choose clean examples for training. \cite{wei2023mitigating} proposes a method called LogitClip to enhance the noise robustness by clamping the norm of the logit vector, ensuring an upper bound at the logit level. \cite{ding2023improve} proposes enhancing robust loss functions by incorporating noise-aware hyperparameters, ensuring improved noise tolerance with a theoretical guarantee. These robust loss functions can somewhat alleviate the overfitting to mislabeled data, but their generalization performance will degrade when complex datasets are involved~\cite{han2020survey}.

\section{Methodology}\label{sec:method}

\subsection{Preliminaries}
We follow \cite{han2020sigua} to define the problem of learning with noisy labels. Specifically, we consider a $k$-class classification task. Let $\mathcal{X}$ and $\mathcal{Y}$ be input and output spaces respectively, where $\mathcal{X}\in\mathbbm{R}^d$, and $\mathcal{Y}=[k]$\footnote{Let $[z]=\{1,2,\ldots,z\}$.}. Denote the random variable pair of interest by $(\mathbf{x},y)$, and $p(\mathbf{x},y)$ be the underlying joint density from which test data will be sampled. In learning with noisy labels, the labels of training data are corrupted before being observed. The training data are sampled from a corrupted joint density $p(\mathbf{x},\tilde{y})$ rather than $p(\mathbf{x},y)$, where $\tilde{y}$ is the random variable of noisy labels. Due to noisy labels, $p(y|\mathbf{x})$ is corrupted into $p(\tilde{y}|\mathbf{x})$. We hence have an observed noisy training sample that is drawn from $p(\mathbf{x},\tilde{y})$, which is denoted by $\{(\mathbf{x}_i,\tilde{y}_i)\}_{i=1}^n$ with a sample size $n$.

Let $f:\mathcal{X}\rightarrow\mathbbm{R}^k$ be the classifier with learnable parameters $\mathbf{w}$. At the $t$-th iteration during
training, the parameters of the classifier can be denoted as $\mathbf{w}_t$. Let $\ell:\mathbbm{R}^k\times\mathcal{Y}\rightarrow\mathbbm{R}$ be a surrogate loss function for classification, \eg, the softmax cross-entropy and mean square loss functions. Without any remedy against noisy labels, the objective function\footnote{For simplicity, we omit the $\ell_1$ regularization term included in the objective function.} to be minimized is $L(\mathbf{w};\mathbf{x},\tilde{y})=\frac{1}{n}\sum_{i=1}^n\ell(f(\mathbf{w};\mathbf{x}_i),\tilde{y}_i)$. The updates  of the parameters $\mathbf{w}$ can be represented
by $\mathbf{w}_{t+1}\leftarrow\mathbf{w}_{t}-\eta\nabla_\mathbf{w}L·$. 
\subsection{Descriptions of Technical Details}
\textbf{Network parameter decomposition.} As discussed above, we decompose the parameters $\mathbf{w}$ into two parts: the parameters for desired memorization and the parameters for undesired memorization. The former corresponds to the memorization of clean data, which is denoted by $\bm{\sigma}$. While, the latter corresponds to the memorization of mislabeled data, which is denoted by $\bm{\gamma}$. Mathematically, we have 
\begin{equation}
    \mathbf{w} = \bm{\sigma} + \bm{\gamma}.
\end{equation}
Recall that the memorization effect of deep networks shows that they would first memorize training data with clean labels and then gradually memorize training data with incorrect labels. Afterward, from the perspective of network parameter updates, $\bm{\sigma}$ would be preferentially updated, and $\bm{\gamma}$ would be updated with a lag. Therefore, if we can strengthen the updates of $\bm{\sigma}$ in early training, and weaken the updates in later training, the desired memorization can be improved, following better robustness brought by $\bm{\sigma}$. On the other hand, for $\bm{\gamma}$, we should limit their updates in early training, and strengthen their updates in later training for undesired memorization. Intuition drives us to design a new objective to achieve our goal.

\noindent\textbf{Objective formulation.} Combining the above analyses, we formulate the objective as 
\begin{equation}\label{eq:objective}
    L_{\text{F}}(\mathbf{w};\mathbf{x},\tilde{y})=L(\mathbf{w};\mathbf{x},\tilde{y})+\beta_1(t)\cdot\Delta\bm{\sigma}+\beta_2(t)\cdot\|\bm{\gamma}\|_2, 
\end{equation}

\begin{figure}[!t]
	\centering
        \includegraphics[width=0.45\textwidth]{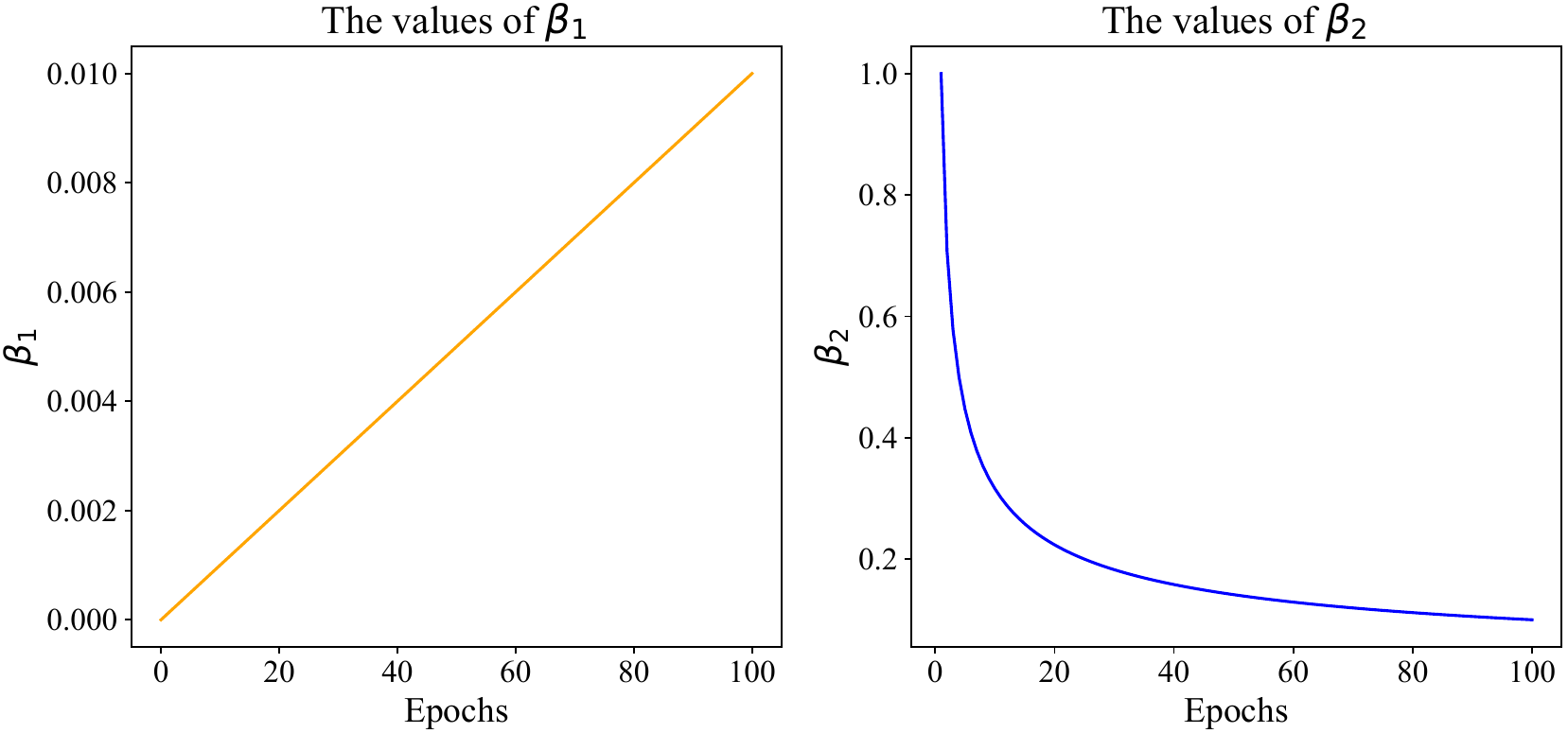}

	\caption{Illustration of the values of $\beta_1$ and $\beta_2$ with varied epochs.}

	\label{fig:beta}
\end{figure}

where $\beta_1(t)$ and $\beta_2(t)$ are two mathematical functions with respect to the epoch $t$, and $\Delta\bm{\sigma}=\|\bm{\sigma} - \bm{\sigma}_{t-1}\|_2$ denotes the Euclidean distance between $\bm{\sigma}$ and $\bm{\sigma}_{t-1}$, where $\bm{\sigma}_{t-1}$ are the parameters $\bm{\sigma}$ obtained in epoch $t-1$. For Eq. (\ref{eq:objective}), the second part controls the updates of $\bm{\sigma}$ with $\beta_1(t)$, and the third part controls the updates of $\bm{\gamma}$ with $\beta_2(t)$. More specifically, $\Delta\bm{\sigma}$ limits the change of parameters between two adjacent epochs. If its weight, \ie, $\beta_1(t)$, is large, $\Delta\bm{\sigma}$ will be reduced fast. Namely, the updates of $\bm{\sigma}$ are restricted. The updates of $\bm{\gamma}$ are bridled in a similar way. 
Note that the constraint forms for $\bm{\sigma}$ and $\bm{\gamma}$ are different. In early training, our goal is to make the memorization of clean data less influenced by $\bm{\gamma}$, ~\ie we want $\bm{\gamma}$ to be as close to zero as possible. $\Delta$ is used to constrain the variation range of the parameter at a given iteration step, while the $\ell_2$ norm is used to constrain the value of the parameter to be as small as possible. Consequently, the $\ell_2$ norm is a more effective constraint form than $\Delta$ in terms of maintaining parameter values at lower magnitudes.

\noindent\textbf{Designs of $\beta_1$ and $\beta_2$.} As analyzed, combining the memorization effect of deep networks, we should dynamically control the updates of $\bm{\sigma}$ and $\bm{\gamma}$. Thereby, in early training, $\bm{\sigma}$ should be updated quickly to fit clean data; in later training, they should be updated slowly to avoid the fitting of mislabeled data. As for $\bm{\gamma}$, the operation on parameter updates is the opposite. 

We hence design $\beta_1$ as an \textit{increasing} function and $\beta_2$ as a \textit{decreasing} function with respect to the epoch $t$. There are many choices of the function types for $\beta_1$ and $\beta_2$, \eg, linear functions, exponential functions, power functions \etal ~In this work, we set $\beta_1(t)=c_1t$ and $\beta_2(t)=t^{-c_2}$ (see Fig.~\ref{fig:beta}), where $c_1$ and $c_2$ are two hyper-parameters. Both the choice and detailed analysis of them will be provided in the experimental sections. 

It is worth noting that prior work~\cite{hu2020simple} discusses that the regularization by the distance between the network parameters to initialization, can make the network robust to noisy labels. Although our method controls the distance of parameters at two adjacent iterations, it is different from prior work~\cite{hu2020simple}. Specifically, in this paper, we have $\|\mathbf{w}-\mathbf{w}_0\|_2=\|\bm{\sigma}-\bm{\sigma}_0+\bm{\gamma}-\bm{\gamma}_0\|_2$. Even though $\|\bm{\sigma}-\bm{\sigma}_0\|_2$ is small, $\|\mathbf{w}-\mathbf{w}_0\|_2$ is not necessary to be small because of $\bm{\gamma}$. The analysis is not contradictory to the claims in~\cite{hu2020simple}, since theoretical results in~\cite{hu2020simple} rely on the deep network with infinite width, which is difficult to establish in practice.

\vspace{3pt}
\begin{algorithm}[h]
    \caption{Procedure of the proposed method.}
    \label{alg:4}
    \begin{algorithmic}[1]
        \STATE \textbf{Input}: Initialization parameters $\mathbf{w}$, noisy training set $\mathcal{D}_t$, noisy validation set $\mathcal{D}_v$, learning rate $\eta$, epoch $t_{max}$, and iteration $I_{max}$;
        \STATE Decompose the parameters $\mathbf{w}$ randomly into two parts: $\bm{\sigma}$ and $\bm{\gamma}$;
        
        \FOR{$t=1,2,...,t_{max}$}
            \STATE \textbf{Shuffle} training set $\mathcal{D}_t$;
            \STATE \textbf{Calculate} $\beta_1(t),\beta_2(t)$ in Eq.~(\ref{eq:objective}); 
            \FOR{$I=1,2,...,I_{max}$}
                \STATE \textbf{Fetch} mini-batch $\mathcal{\overline{D}}_t$ from $\mathcal{D}_t$;
                \STATE \textbf{Obtain} $L_{\text{F}}(\mathbf{w};\mathcal{\overline{D}}_t)$ by Eq.~(\ref{eq:objective});
                \STATE \textbf{Update} $\bm{\sigma}=\bm{\sigma}-\eta\nabla_{\bm{\sigma}}L_{\text{F}}(\mathbf{w};\mathcal{\overline{D}}_t)$;
                \STATE \textbf{Update} $\bm{\gamma}=\bm{\gamma}-\eta\nabla_{\bm{\gamma}}L_{\text{F}}(\mathbf{w};\mathcal{\overline{D}}_t)$;
            \ENDFOR
        \ENDFOR
        \STATE // The training is stopped when $\bm{\sigma}$ reach the minimum classification error on the validation set $\mathcal{D}_v$.
        \STATE \textbf{Output}: Parameters $\bm{\sigma}$ after update for predictions.
        
    \end{algorithmic}
\end{algorithm}

\begin{figure*}[!t]
	\centering
        \includegraphics[width=1.0\textwidth]{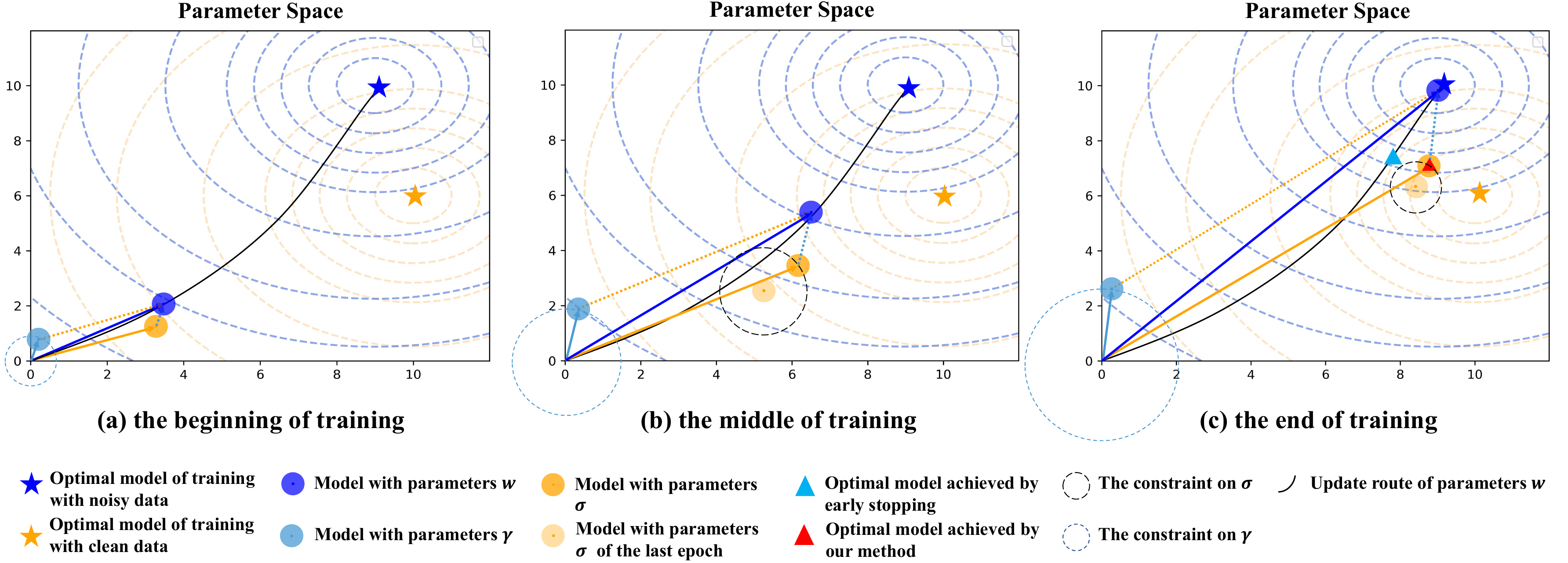}
        \vspace{1pt}
	\caption{Illustrations of justifying our method from the perspective of parameter space.}
	\label{fig:parameterspace}
\end{figure*}

\noindent\textbf{Algorithm flow.} In the training procedure, given noisy training data $\{(\mathbf{x}_i,\tilde{y}_i)\}_{i=1}^n$, we train a deep network with parameters $\mathbf{w}$ on it with Eq.~(\ref{eq:objective}). Both the parameters $\bm{\sigma}$ and $\bm{\gamma}$ are updated by gradient descent. In the test procedure, we only exploit the learned parameters $\bm{\sigma}$ to infer proper labels for given instances. As shown by the algorithm flow, the implementation of our method is convenient, making it easy to apply in practice. 

\subsection{Justification from Parameter Space}
To help further understand the mechanism of our method, we illustrate the training process of three kinds of parameters (\ie, $\mathbf{w}$, $\bm{\sigma}$, and $\bm{\gamma}$) in terms of parameter space in Fig.~\ref{fig:parameterspace}. The three subfigures from left to right represent three training stages respectively. They are related to the beginning of training, the middle of training, and the end of training. 

First, we show the training process of the deep model with parameters $\mathbf{w}$. The yellow star denotes the optimal model of training with clean data while the blue star denotes the optimal model of training with noisy data. The black curve represents the update route of $\mathbf{w}$. As the black curve shows, in the early training step, affected by the memorization effect, $\mathbf{w}$ learn more clean data. The update direction of $\mathbf{w}$ hence moves towards the yellow star. In the later training step, due to the side effect of mislabeled data, the update direction of $\mathbf{w}$ is gradually biased towards the blue star.

Then, we analyze the training process of $\bm{\sigma}$ and $\bm{\gamma}$. As shown in Fig.~\ref{fig:parameterspace}, our method decomposes $\mathbf{w}$ into $\bm{\sigma}$ (shown as the yellow circle) and $\bm{\gamma}$ (shown as the light blue circle). Therefore, the parameters $\mathbf{w}$ can be seen as the sum of two parameter vectors. We introduce a constraint on $\bm{\sigma}$ that increases with epoch $t$, so the black dotted circle in Fig.~\ref{fig:parameterspace} shrinks as the constraint on $\bm{\sigma}$ increases. In the early training stage, we encourage $\bm{\sigma}$ to be updated toward the yellow star. In the later training stage, we limit the updates of $\bm{\sigma}$ to avoid $\bm{\sigma}$ being disturbed by noise and thus deviating from the yellow star. For $\bm{\gamma}$, we impose a decreasing constraint, so the blue dotted circle keeps expanding as the constraint on $\bm{\gamma}$ decreases. In early training, the updates of $\bm{\gamma}$ are restricted, and clean data can be fully learned by $\bm{\sigma}$. As the training goes on, $\mathbf{w}$ will be biased towards the blue star. At this time, we encourage $\bm{\gamma}$ to be updated to reduce the impact of bias on $\bm{\sigma}$.

Finally, we analyze the superiority of our method. As we can see, the blue triangle is the optimal model trained by early stopping because this is the closest point to the yellow star in the training trajectory. Benefiting from the proposed parameter decomposition and effective constraints, the best model with $\bm{\sigma}$ (shown as the red triangle) that we use in the inference time is closer to the yellow star. This intuitively explains the superiority of the proposed method.


\section{Experiments}\label{sec:experiments}
In this section, we exploit comprehensive experiments to verify the effectiveness of our method.

\subsection{Datasets and Implementation Details}\label{sec:4.1}
\textbf{Simulated noisy datasets.} We validate our method on four simulated noisy datasets, \ie, simulated MNIST~\cite{LeCunmnist}, F-MNIST~\cite{xiao2017fashion}, CIFAR-10~\cite{krizhevsky2009learning}, and CIFAR-100~\cite{krizhevsky2009learning}. The four datasets are popularly used to verify algorithm robustness against noisy labels~\cite{ma2020normalized,li2022selective,liu2020early,zhou2021learning,kim2019nlnl,mirzasoleiman2020coresets,wang2022scalable,tanno2019learning,han2019deep}.

\begin{table}[!t]
    \centering
    \scriptsize
    \vspace{3pt}
    \renewcommand{\arraystretch}{1.4}
    \begin{tabular}{l|c|c|c|c}
    \hline
    Datasets &\# of training & \# of testing & \# of classes & Size  \\\hline
    MNIST     & 60,000 & 10,000 & 10 & 28$\times$28$\times$1\\ \hline
    F-MNIST & 60,000 & 10,000 & 10 & 28$\times$28$\times$1\\\hline
    CIFAR-10 & 50,000 & 10,000 & 10 & 32$\times$32$\times$3\\\hline
    CIFAR-100 & 50,000 & 10,000 & 100 & 32$\times$32$\times$3\\\hline
    Food-101 & 75,750 & 25,250 & 101 & 112$\times$112$\times$3 \\\hline
    Clothing1M & 1,000,000 & 10,000 & 14 & 224$\times$224$\times$3\\\hline
    CIFAR-10N & 50,000 & 10,000 & 10 & 32$\times$32$\times$3\\\hline
    \end{tabular}
    \vspace{3pt}
    \caption{Summary of used noisy datasets in experiments.}
    \label{tab:important_statistics}
\end{table}

In this paper, we consider four kinds of simulated noises to generate noisy labels, which are (1)  Symmetric noise (abbreviated as Sym.): Each class has the same probability of incorrect flipping to any other class. (2) Asymmetric noise (abbreviated as Asym.): Similar classes are mistakenly flipped between each other. Specifically, we perform ``2$\rightarrow$7, 5$\leftrightarrow$6, 3$\rightarrow$8'' for MNIST, ``Pullover$\rightarrow$Coat, Sandals$\rightarrow$Sneaker, T-shirt$\rightarrow$Shirt'' for F-MNIST, and  ``Bird$\rightarrow$Airplane, Deer$\rightarrow$Horse, Truck$\rightarrow$Automobile, Cat$\leftrightarrow$Dog'' for CIFAR-10. Lastly, for CIFAR-100, all classes
are grouped into 20 super-classes, and each has 5 sub-classes. Each class is then flipped into the next within the same super-class. (3) Pairflip noise (abbreviated as Pair.): Each class can only be incorrectly flipped to its neighboring classes. (4) Instance-dependent noise (abbreviated as Ins.): The generation of such noise is affected by image features, which is more consistent with the noise generation process in the real world, and more challenging. We follow previous works~\cite{xia2020part,zhu2021second} for instance-dependent noise generation. For all kinds of noise, the noise rate is set to 20\% and 40\%. 10\% noisy training data is employed for validation.

We exploit a LeNet-5 for MNIST, a ResNet-18 for F-MNIST and CIFAR-10, and a ResNet-50 for CIFAR-100. Data augmentations including random cropping and horizontal flipping are used for  CIFAR-10 and CIFAR-100. In addition, an SGD optimizer with 0.9 momentum and a weight decay of 0.001 is employed. Batch size is set to 32 for MNIST and F-MNIST, and 64 for CIFAR-10 and CIFAR-100. Besides, the initial learning rate is 0.01, which is decayed by 0.1 at the 10th and 20th epochs for MNIST and F-MNIST, and at the 40th and 80th epochs for CIFAR-10 and CIFAR-100. 100 epochs are set totally. We set $c_1=10^{-4}$ consistently while tuning $c_2$ with the noisy validation set. 

\begin{table*}[!t]
	\renewcommand\arraystretch{1.2}
	\caption{Mean and standard deviation of test accuracy (\%) on simulated noisy datasets under different noise settings. Experimental results are reported over five trials. Best experimental results are \textbf{boldfaced}.}

	\begin{center}
                \begin{tabular}{l|l|cc|cc|cc|cc}
                \hline
                & Method & Sym.~20\% & Sym.~40\% & Asym.~20\% & Asym.~40\% & Pair.~20\% & Pair.~40\% & Ins.~20\% & Ins.~40\%  \\
                \hline
                \multirow{9}{*}{\rotatebox[origin=c]{90}{MNIST}} & Standard & 98.95$\pm$0.12 & 98.75$\pm$0.10 & 99.20$\pm$0.02 & 98.13$\pm$0.57 & 99.06$\pm$0.05 & 97.83$\pm$0.41 & 98.51$\pm$0.21 & 93.34$\pm$1.63\\
                
                & CoTeaching & 98.77$\pm$0.08 & 98.41$\pm$0.09 & 98.83$\pm$0.05 & 97.85$\pm$0.11 & 98.83$\pm$0.13 & 97.89$\pm$0.22 & 98.73$\pm$0.06 & 98.13$\pm$0.24\\

                & CoTeaching+ & 98.82$\pm$0.21 & 98.62$\pm$0.08 & 98.97$\pm$0.14 & 93.96$\pm$5.39 & 98.85$\pm$0.07 & 96.06$\pm$4.40 & \textbf{98.93}$\pm$\textbf{0.06} & \textbf{98.46}$\pm$\textbf{0.26}\\

                & JoCor & 98.66$\pm$0.14 & 98.41$\pm$0.11 & 98.98$\pm$0.07 & 98.16$\pm$0.17 & 98.79$\pm$0.10 & 97.94$\pm$0.31 & 98.82$\pm$0.08 & 98.17$\pm$0.11 \\

                & SIGUA & 98.44$\pm$0.17 & 98.03$\pm$0.21 & 97.20$\pm$0.49 & 95.06$\pm$1.33 & 98.45$\pm$0.13 & 96.25$\pm$1.23 & 98.49$\pm$0.16 & 96.91$\pm$0.47 \\

                & CNLCU & 98.71$\pm$0.11 & 98.39$\pm$0.13 & 99.07$\pm$0.10 & 98.55$\pm$0.35 & 98.79$\pm$0.09 & 98.16$\pm$0.39 & 98.53$\pm$0.11 & 97.59$\pm$0.16 \\ 

                & AdaCorr & 98.97$\pm$0.06 & 98.63$\pm$0.11 & 99.01$\pm$0.12 & 98.48$\pm$0.13 & 98.88$\pm$0.14 & 98.10$\pm$0.12 & 98.62$\pm$0.15 & 97.12$\pm$0.27 \\
                
                & CDR & 98.99$\pm$0.07 & 98.79$\pm$0.06 & 99.19$\pm$0.06 & 98.27$\pm$0.53 & 99.12$\pm$0.04 & 97.53$\pm$0.32 & 98.51$\pm$0.08 & 93.36$\pm$1.10 \\\cline{2-10}
                
                & TNLPAD (ours) & \textbf{99.02}$\pm$\textbf{0.07} & \textbf{98.84}$\pm$\textbf{0.06} & \textbf{99.30}$\pm$\textbf{0.03} & \textbf{98.97}$\pm$\textbf{0.16} & \textbf{99.17}$\pm$\textbf{0.03} & \textbf{98.75}$\pm$\textbf{0.11} & 98.84$\pm$0.07 & 97.01$\pm$0.38 \\
                \hline\hline

                \multirow{9}{*}{\rotatebox[origin=c]{90}{F-MNIST}} & Standard & 91.87$\pm$0.35 & 90.58$\pm$0.23 & 92.68$\pm$0.16 & 89.32$\pm$0.65 & 93.21$\pm$0.15 & 86.42$\pm$1.91 & 91.47$\pm$0.19 & 86.19$\pm$0.72\\
                
                & CoTeaching & 92.28$\pm$0.23 & 90.86$\pm$0.12 & 93.10$\pm$0.31 & 87.14$\pm$2.72 & 92.78$\pm$0.40 & 91.44$\pm$0.27 & 92.32$\pm$0.18 & 89.96$\pm$0.88\\
	
	            & CoTeaching+ & 92.60$\pm$0.29 & 91.34$\pm$0.43 & 92.74$\pm$0.22 & 67.18$\pm$0.18 & 93.05$\pm$0.27 & 80.11$\pm$4.10 & 92.66$\pm$0.19 & 66.92$\pm$6.52\\

                & JoCor & 92.11$\pm$0.10 & 90.62$\pm$0.25 & 92.65$\pm$0.21 & 88.99$\pm$0.72 & 92.87$\pm$0.35 & 90.95$\pm$0.42 & 92.34$\pm$0.18 & 89.53$\pm$0.31 \\

                & SIGUA & 91.55$\pm$0.55 & 90.13$\pm$0.42 & 89.19$\pm$0.22 & 76.94$\pm$4.54 & 92.12$\pm$0.58 & 89.18$\pm$0.56 & 91.68$\pm$0.14 & 88.17$\pm$1.75\\

                & CNLCU & 91.85$\pm$0.24 & 90.62$\pm$0.39 & 93.02$\pm$0.13 & 90.36$\pm$0.82 & 92.76$\pm$0.21 & 90.99$\pm$1.10 & 91.96$\pm$0.41 & 88.84$\pm$0.48 \\
                
                & AdaCorr & 90.38$\pm$0.53 & 88.73$\pm$0.73 & 92.38$\pm$0.24 & 88.81$\pm$0.57 & 93.02$\pm$0.28 & 89.51$\pm$0.31 & 91.46$\pm$0.52 & 85.50$\pm$1.42\\

                & CDR & 91.84$\pm$0.17 & 90.38$\pm$0.41 & 92.59$\pm$0.45 & 88.91$\pm$0.53 & 93.10$\pm$0.18 & 87.46$\pm$3.11 & 91.34$\pm$0.16 & 82.24$\pm$2.66\\\cline{2-10}
                
                & TNLPAD (ours) & \textbf{93.11}$\pm$\textbf{0.23} & \textbf{91.98}$\pm$\textbf{0.29} & \textbf{93.91}$\pm$\textbf{0.05} & \textbf{91.95}$\pm$\textbf{0.29} & \textbf{94.10}$\pm$\textbf{0.07} & \textbf{92.96}$\pm$\textbf{0.39} & \textbf{93.22}$\pm$\textbf{0.25} & \textbf{90.16}$\pm$\textbf{0.80}\\

                \hline\hline
                
                \multirow{9}{*}{\rotatebox[origin=c]{90}{CIFAR-10}} & Standard & 90.19$\pm$0.31 & 86.26$\pm$0.11 & 91.49$\pm$0.60 & 87.77$\pm$1.15 & 91.56$\pm$0.13 & 87.36$\pm$1.04 & 90.57$\pm$0.19 & 83.99$\pm$0.37\\

                & CoTeaching & 88.52$\pm$0.14 & 85.53$\pm$0.78 & 90.32$\pm$0.14 & 78.40$\pm$0.27 & 88.87$\pm$0.32 & 83.52$\pm$0.44 & 88.51$\pm$0.24 & 83.82$\pm$1.10 \\

                & CoTeaching+ & 88.87$\pm$0.07 & 85.71$\pm$0.44 & 88.48$\pm$0.18 & 66.40$\pm$4.03 & 88.76$\pm$0.30 & 80.54$\pm$3.76 & 88.52$\pm$0.12 & 81.91$\pm$4.81  \\

                & JoCor & 89.08$\pm$0.20 & 85.37$\pm$0.41 & 89.64$\pm$0.21 & 84.23$\pm$0.33  & 88.70$\pm$0.36 & 81.29$\pm$0.97 & 88.62$\pm$0.44 & 82.28$\pm$1.51   \\

                & SIGUA & 67.60$\pm$26.73 & 81.28$\pm$0.75 & 85.07$\pm$0.25 & 59.39$\pm$1.83 & 86.17$\pm$0.64 & 68.08$\pm$3.49 & 86.55$\pm$0.79 & 75.38$\pm$2.19\\

                & CNLCU &  87.53$\pm$0.25 & 84.49$\pm$0.39 & 90.18$\pm$0.14 & 83.70$\pm$1.37 & 88.69$\pm$0.29 & 83.88$\pm$0.61 & 88.05$\pm$0.35 & 83.39$\pm$0.48\\
                               
                & AdaCorr & 84.02$\pm$0.28 & 79.22$\pm$0.57 & 85.31$\pm$0.37 & 80.86$\pm$0.89 & 85.98$\pm$0.37 & 76.48$\pm$0.34 & 85.22$\pm$0.56 & 77.04$\pm$0.42\\

                & CDR & 89.84$\pm$0.31 & 86.30$\pm$0.45 & 91.62$\pm$0.38 & 88.42$\pm$0.85 & 91.59$\pm$0.35 & 87.14$\pm$0.92 & 90.20$\pm$0.53 & 82.65$\pm$0.74\\\cline{2-10}
                
                & TNLPAD (ours) & \textbf{91.07}$\pm$\textbf{0.19} & \textbf{87.83}$\pm$\textbf{0.31} & \textbf{92.17}$\pm$\textbf{0.13} & \textbf{89.25}$\pm$\textbf{0.60} & \textbf{92.18}$\pm$\textbf{0.19} & \textbf{89.06}$\pm$\textbf{0.71} & \textbf{91.64}$\pm$\textbf{0.12} & \textbf{87.39}$\pm$\textbf{0.45}\\
                \hline\hline

                \multirow9{*}{\rotatebox[origin=c]{90}{CIFAR-100}} & Standard & 65.77$\pm$0.61 & 60.58$\pm$0.34 & 66.55$\pm$0.74 & 52.51$\pm$0.53 & 66.43$\pm$0.48 & 52.92$\pm$0.85 & 65.01$\pm$0.51 & 57.88$\pm$0.73\\
                
                & CoTeaching & 61.64$\pm$0.40 & 54.59$\pm$1.24 & 57.92$\pm$0.57 & 41.42$\pm$0.38 & 57.79$\pm$0.53 & 40.24$\pm$1.35 & 60.14$\pm$0.46 & 48.81$\pm$0.92 \\

                & CoTeaching+ & 61.75$\pm$1.06 & 43.85$\pm$2.07 & 61.43$\pm$1.07 & 41.98$\pm$2.10 & 61.26$\pm$1.47 & 43.04$\pm$1.19 & 60.43$\pm$1.48 & 45.56$\pm$3.71\\

                & JoCor & 58.07$\pm$3.43 & 46.81$\pm$2.91 & 54.73$\pm$1.49 & 35.55$\pm$0.65 & 53.36$\pm$3.38 & 33.33$\pm$1.55 & 53.56$\pm$1.53 & 40.57$\pm$1.38 \\

                & SIGUA & 56.28$\pm$2.14 & 50.16$\pm$1.21 & 53.96$\pm$2.35 & 39.08$\pm$1.30 & 54.17$\pm$0.96 & 38.64$\pm$0.80 & 54.50$\pm$1.53 & 44.08$\pm$2.06 \\

                & CNLCU & 61.30$\pm$0.66 & 54.87$\pm$0.80 & 56.84$\pm$0.81 & 43.48$\pm$1.08 & 57.69$\pm$0.64 & 42.57$\pm$1.23 & 58.37$\pm$0.67 & 49.32$\pm$0.96 \\
                                
                & AdaCorr & 55.78$\pm$0.48 & 49.75$\pm$0.85 & 55.55$\pm$0.72 & 41.23$\pm$1.24 & 54.84$\pm$1.00 & 41.75$\pm$1.21 & 55.17$\pm$0.84 & 45.12$\pm$0.99\\

                & CDR & 67.96$\pm$0.25 & 61.79$\pm$0.33 & 70.07$\pm$0.41 & 54.58$\pm$1.24 & 71.44$\pm$0.60 & 55.37$\pm$0.84 & 70.15$\pm$0.14 & 61.56$\pm$0.42\\\cline{2-10}
                
                & TNLPAD (ours) & \textbf{69.46}$\pm$\textbf{0.25} & \textbf{63.63}$\pm$\textbf{0.38} & \textbf{72.17}$\pm$\textbf{0.68} & \textbf{57.21}$\pm$\textbf{1.25} & \textbf{72.13}$\pm$\textbf{0.36} & \textbf{58.16}$\pm$\textbf{1.29} & \textbf{71.05}$\pm$\textbf{0.83} & \textbf{63.03}$\pm$\textbf{0.99}\\
                \hline
                \end{tabular}
	\end{center}
	\label{tab:simulated_experiments}
\end{table*}

\begin{table*}[!t]
    \centering
    \small
    \vspace{2pt}
    \renewcommand{\arraystretch}{1.2}  
    \caption{Test accuracy (\%) on real-world noisy datasets. Best experimental results are \textbf{boldfaced}.}
    \begin{tabular}{l|c|c|cccc}
    \hline
         Methods &  Food-101 & Clothing1M & CIFAR-10N-1 & CIFAR-10N-2 & CIFAR-10N-3 & CIFAR-10N-W\\\hline
         Standard& 73.63 & 68.98 & 88.31$\pm$0.60  & 88.11$\pm$0.16 & 88.17$\pm$0.15 & 80.10$\pm$0.43\ \\
         CoTeaching& 71.26 & 68.91 &  89.49$\pm$0.22  & 89.29$\pm$0.13 & 89.25$\pm$0.24 & 78.86$\pm$0.97 \\
         CoTeaching+& 73.27 & 68.44 &  88.20$\pm$0.29 & 88.32$\pm$0.16 & 88.20$\pm$ 0.30 & 81.05$\pm$0.04 \\
         JoCor& 66.48 & 69.62 &  89.29$\pm$0.19  & 89.28$\pm$0.32 & 88.84$\pm$0.17 & 78.91$\pm$0.69 \\
         SIGUA& 66.38 & 65.00 & 86.79$\pm$2.04  & 87.58$\pm$1.00 & 85.77$\pm$1.95 & 76.79$\pm$0.73 \\
         CNLCU& 71.26 & 68.13 & 87.02$\pm$0.26  & 86.37$\pm$0.21 & 86.38$\pm$0.30 & 77.49$\pm$0.54 \\
         AdaCorr& 74.56 & 65.56 & 87.81$\pm$0.33  & 87.95$\pm$0.26 & 88.36$\pm$ 0.40 & 82.19$\pm$0.77\\
         CDR & 72.67 & 68.11 & 88.20$\pm$0.44  & 88.28$\pm$0.25 & 88.17$\pm$0.46 & 80.28$\pm$0.68\\\hline
         TNLPAD (ours)& \textbf{75.97} & \textbf{71.51} & \textbf{89.95}$\pm$\textbf{0.34}  & \textbf{89.55}$\pm$\textbf{0.30} & \textbf{89.70}$\pm$\textbf{0.22} & \textbf{83.84}$\pm$\textbf{0.35} \\\hline 
    \end{tabular}
    \label{tab:real_world}
\end{table*}

\begin{table}[!t]
    \centering
    \vspace{3pt}
    \renewcommand{\arraystretch}{1.25}
    \caption{Results of combining the proposed method with semi-supervised learning. ``A*'' means that the method ``A'' is boosted by our method. Best experimental results are \textbf{boldfaced}.}
    \begin{tabular}{l|cccc}
    \hline
    {Method} & Sym.~20\% & Sym.~40\% & Asym.~20\% & Asym.~40\% \\\cline{1-5}
    Self &  92.65$\pm$0.16 & 88.67$\pm$0.03 & 91.43$\pm$0.52 & 86.22$\pm$0.13 \\
    Self* & \textbf{92.78$\pm$0.61} & \textbf{89.47$\pm$0.24} & \textbf{92.26$\pm$0.41} & \textbf{87.06$\pm$0.34}\\\cline{1-5}
    DivideMix & \textbf{96.10$\pm$0.14} & 94.50$\pm$0.26 & 94.47$\pm$0.11 & 92.16$\pm$0.32\\
    DivideMix* & 95.99$\pm$0.03 & \textbf{94.66$\pm$0.11} & \textbf{94.73$\pm$0.09} & \textbf{92.54$\pm$0.19}\\\hline
    \end{tabular}
    \label{tab:combination_with_semi}
\end{table}

\noindent\textbf{Real-world noisy datasets.} We validate our method on three real-world noisy datasets, \ie, Food-101~\cite{bossard2014food}, Clothing1M~\cite{xiao2015learning}, and CIFAR-10N~\cite{wei2022learning}. The important statistics of all used datasets are provided in Table~\ref{tab:important_statistics}. Additionally, for Food-101, a ResNet-50 (not pre-trained) is exploited. We use an SGD optimizer with 0.9 momentum, batch size 32, and the initial learning rate 0.001 to train the network. For Clothing1M, we exploit a ResNet-50 pre-trained on ImageNet. We use an SGD optimizer with 0.9 momentum. The weight decay and batch size are adjusted to 0.005 and 32. The initial learning rate is set to 0.001 and then divided by 10 after the 5th epoch. The maximum number of epochs is set to 20. Limited by computational cost, we only conduct one experiment on Food-101 and Clothing1M. For CIFAR-10N, we exploit a ResNet-18 and we also use SGD with 0.9 momentum. The weight decay and batch size are set to 0.001 and 64. The initial learning rate is set to 0.01, which is decayed by 0.1 at the 40th epoch and the 80th epoch respectively.

\begin{table}[!h]
    \centering
    \vspace{3pt}
    \renewcommand{\arraystretch}{1.2}
    \caption{Mean and standard deviation of test accuracy (\%) on simulated noisy datasets with high noise levels (60\%, 70\%, and 80\%). Experimental results are reported over five trials. Best experimental results are \textbf{boldfaced}.}
    \begin{tabular}{l|l|ccc}
    \hline
    & Method & Sym.~60\% & Sym.~70\% & Sym.~80\% \\\hline
   \multirow{9}{*}{\rotatebox[origin=c]{90}{CIFAR-10}} & Standard & 78.33$\pm$3.46 & 71.62$\pm$1.21 & 48.89$\pm$1.78\\ 
    & CoTeaching & 77.32$\pm$1.03 & 68.58$\pm$1.03 & 25.97$\pm$1.68 \\
    & CoTeaching+ & 72.18$\pm$4.19 & 41.24$\pm$2.29 & 18.07$\pm$3.77\\
    & JoCor& 76.31$\pm$0.92 & 55.31$\pm$3.83 & 22.13$\pm$4.16\\
    & SIGUA & 72.67$\pm$0.53 &25.06$\pm$3.04 & 18.82$\pm$1.87\\
    & CNLCU & 76.13$\pm$0.83 & 65.30$\pm$2.00 & 19.63$\pm$1.30\\
    & AdaCorr & 71.42$\pm$0.86 & 62.75$\pm$1.25 & 41.84$\pm$1.34 \\
    & CDR & 79.20$\pm$0.31 & 70.10$\pm$1.04 & 47.15$\pm$4.24\\\cline{2-5}
    & TNLPAD (ours) & \textbf{79.95}$\pm$\textbf{1.28} & \textbf{72.00}$\pm$\textbf{1.46} &\textbf{49.23}$\pm$\textbf{1.64}\\
    \hline\hline
    \multirow{9}{*}{\rotatebox[origin=c]{90}{CIFAR-100}} & Standard & 48.62$\pm$2.94 & 38.12$\pm$1.13 & 21.15$\pm$0.85\\ 
    & CoTeaching & 42.92$\pm$1.73 & 31.44$\pm$2.23 & 14.80$\pm$1.35\\
    & CoTeaching+ & 12.80$\pm$3.81 & 6.73$\pm$0.42 & 3.48$\pm$0.29\\
    & JoCor & 33.00$\pm$1.61 & 20.96$\pm$2.15 & 7.09$\pm$0.72\\
    & SIGUA & 29.77$\pm$3.31 & 19.71$\pm$2.67 & 2.45$\pm$0.82\\
    & CNLCU & 41.01$\pm$2.47 & 19.12$\pm$1.75 & 3.85$\pm$0.47\\
    & AdaCorr & 38.79$\pm$1.36 & 29.73$\pm$1.21 & 18.80$\pm$1.38\\
    & CDR & 51.37$\pm$1.27 & 39.34$\pm$0.80 & 21.28$\pm$0.80\\ \cline{2-5}
    & TNLPAD (ours) & \textbf{51.54}$\pm$\textbf{0.85} & \textbf{39.51}$\pm$\textbf{1.18} & \textbf{22.28}$\pm$\textbf{0.30}\\\hline
    \end{tabular}
    \label{tab:high_noise_levels}
\end{table}

\begin{table*}[!t]
    \centering
    \small
    \renewcommand{\arraystretch}{1.2}
    \caption{Mean and standard deviation of test accuracy (\%) on simulated noisy datasets with multiple network architectures. Experimental results are reported over five trials. Best experimental results are \textbf{boldfaced}.}
    \begin{tabular}{l|l|cc|cc|cc}
    \hline
    &\multirow{2}{*}{Method}&\multicolumn{2}{c|}{Wide-ResNet-28} & \multicolumn{2}{c|}{VGG-19} & \multicolumn{2}{c}{ResNeXt-29}\\\cline{3-8}
    &  & Sym.~20\% & Sym.~40\% & Sym.~20\% & Sym.~40\% &  Sym.~20\% & Sym.~40\% \\\hline
   \multirow{9}{*}{\rotatebox[origin=c]{90}{CIFAR-10}} & Standard & 89.95$\pm$0.26 & 85.58$\pm$0.17 & 88.98$\pm$0.32 & 85.71$\pm$0.28 & 89.81$\pm$0.25 & 85.47$\pm$0.54\\ 
    & CoTeaching & 89.21$\pm$0.38 & 85.22$\pm$0.69  & 88.45$\pm$0.17 & 84.85$\pm$0.35 & 87.70$\pm$0.38 & 83.61$\pm$0.38\\
    & CoTeaching+ & 90.38$\pm$0.36 & 86.77$\pm$0.38 & 86.38$\pm$0.50 & 59.39$\pm$6.61 & 87.99$\pm$0.31 & 84.13$\pm$0.81\\
    & JoCor & 88.78$\pm$0.35 & 85.32$\pm$0.78 & 87.46$\pm$0.67 & 68.12$\pm$11.38 & 87.99$\pm$0.54 & 84.29$\pm$0.54\\
    & SIGUA & 88.43$\pm$0.61 & 83.35$\pm$0.81 & 63.80$\pm$7.35 & 25.69$\pm$7.08 & 86.43$\pm$0.66 & 81.40$\pm$1.69\\
    & CNLCU & 87.13$\pm$0.42 & 83.52$\pm$0.54 & 87.48$\pm$0.37 & 82.59$\pm$2.29 & 86.82$\pm$0.04 & 83.03$\pm$0.47\\
    & AdaCorr & 84.71$\pm$0.59 & 80.19$\pm$0.95 & 83.96$\pm$0.49 & 79.44$\pm$1.07 & 82.39$\pm$0.62 & 76.88$\pm$1.00\\
    & CDR & 89.94$\pm$0.17 & 85.70$\pm$0.32 & 89.26$\pm$0.20 & 85.99$\pm$0.54 & 90.26$\pm$0.13 & 86.61$\pm$0.08\\\cline{2-8}
    & TNLPAD (ours) & \textbf{91.83$\pm$0.22} & \textbf{88.57$\pm$0.66} & \textbf{89.55$\pm$0.22} & \textbf{86.21$\pm$0.27} & \textbf{90.65$\pm$0.28} & \textbf{86.94$\pm$0.29}\\\hline\hline
    \multirow{9}{*}{\rotatebox[origin=c]{90}{CIFAR-100}} & Standard & 66.86$\pm$0.57 & 57.80$\pm$0.24 &60.36$\pm$0.32 & 50.60$\pm$0.28 & 65.98$\pm$0.28 & 58.13$\pm$0.62\\ 
    & CoTeaching & 63.36$\pm$0.62 & 57.06$\pm$0.60 & 34.13$\pm$3.73 & 16.83$\pm$2.40 & 61.53$\pm$0.37 & 55.92$\pm$0.34\\
    & CoTeaching+ & 64.07$\pm$0.80 & 49.71$\pm$1.55 & 48.58$\pm$3.42 & 24.45$\pm$4.85 & 63.17$\pm$0.79 & 51.50$\pm$0.53\\
    & JoCor& 64.00$\pm$0.50 & 56.99$\pm$0.71 & 28.54$\pm$3.38 & 8.74$\pm$1.14 & 61.81$\pm$0.81 & 55.42$\pm$0.34\\
    & SIGUA & 63.42$\pm$0.77 & 54.93$\pm$1.92 & 14.44$\pm$5.86 & 5.16$\pm$0.64 & 60.87$\pm$0.97 & 53.56$\pm$0.56\\
    & CNLCU & 62.41$\pm$0.47 & 56.64$\pm$0.16 & 51.58$\pm$4.28 & 29.04$\pm$3.67 & 61.35$\pm$0.51 & 56.59$\pm$0.28\\
    & AdaCorr & 56.32$\pm$0.71 & 49.32$\pm$0.48 & 42.63$\pm$0.84 & 29.77$\pm$1.18 & 57.35$\pm$0.36 & 49.23$\pm$0.82\\
    & CDR & 68.45$\pm$0.55 & 61.71$\pm$0.51 & 60.51$\pm$0.25 & \textbf{53.50$\pm$1.10} & 68.66$\pm$0.15 & 62.50$\pm$0.40\\ \cline{2-8}
    & TNLPAD (ours) & \textbf{71.39$\pm$0.23} & \textbf{65.41$\pm$0.44} & \textbf{61.19$\pm$0.28} & 52.34$\pm$0.30 & \textbf{70.87$\pm$0.20} & \textbf{64.66$\pm$0.79}\\\hline
    \end{tabular}
    \label{tab:networks}
    \vspace{3pt}
\end{table*}

\begin{figure*}[h]
    \centering
    \subfloat[]{
		\label{fig:subfig:decre_function} 
        \includegraphics[width=0.33\linewidth, height=4.5cm]{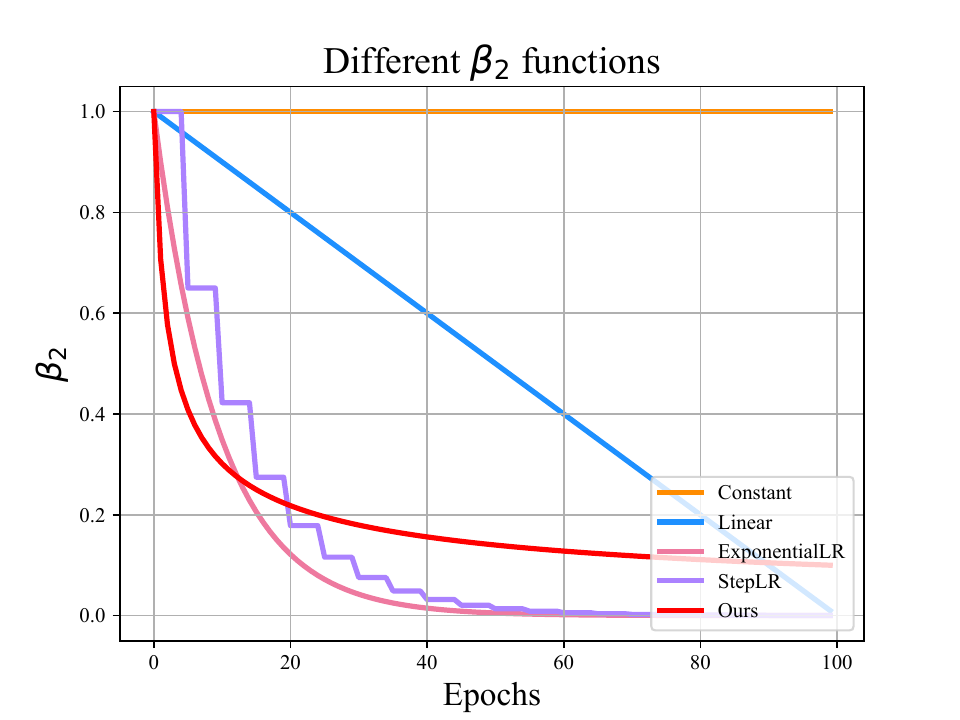}}
    \subfloat[]{
		\label{fig:subfig:fmnist-symmetric-20} 
        \includegraphics[width=0.33\linewidth, height=4.5cm]{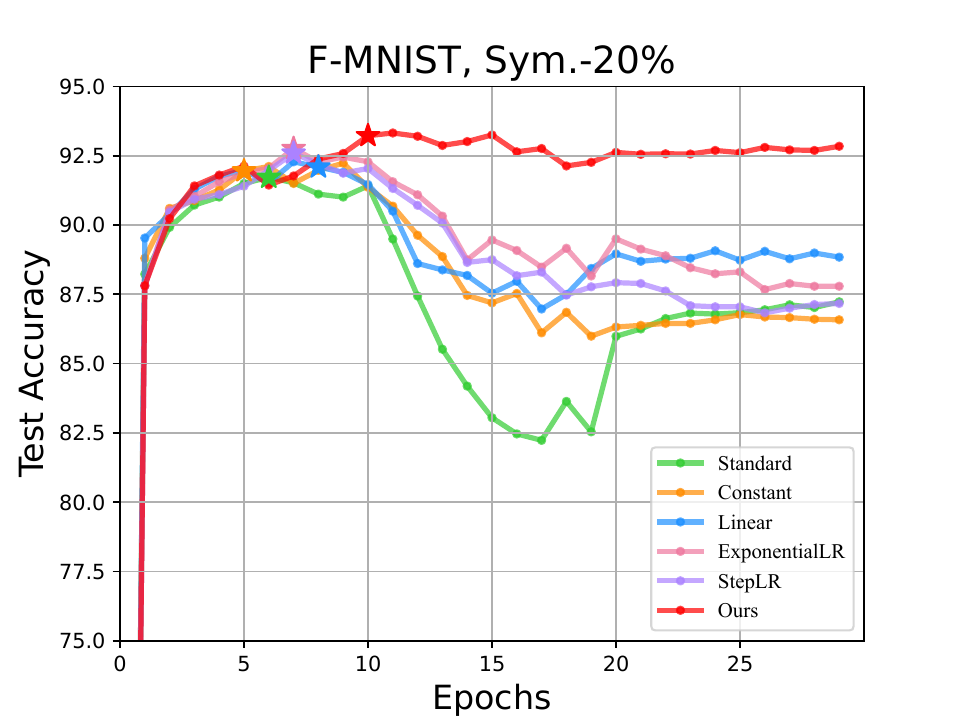}}
    \subfloat[]{
		\label{fig:subfig:fmnist-symmetric-40} 
		\includegraphics[width=0.33\linewidth, height=4.5cm]{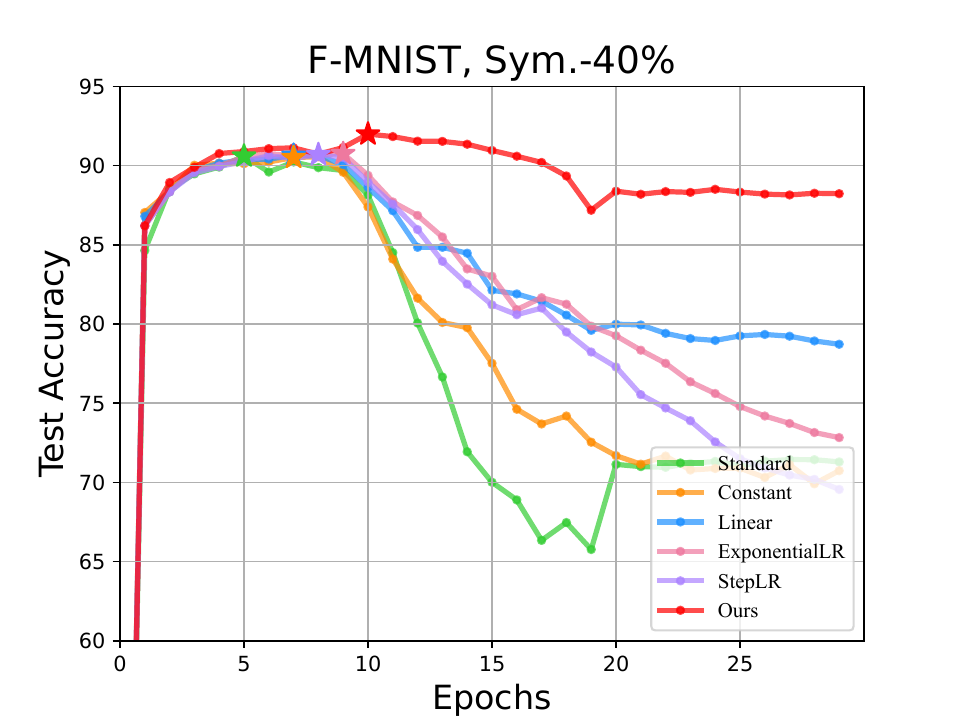}}

    \subfloat[]{
		\label{fig:subfig:cifar10-symmetric-20} 
		\includegraphics[width=0.33\linewidth, height=4.5cm]{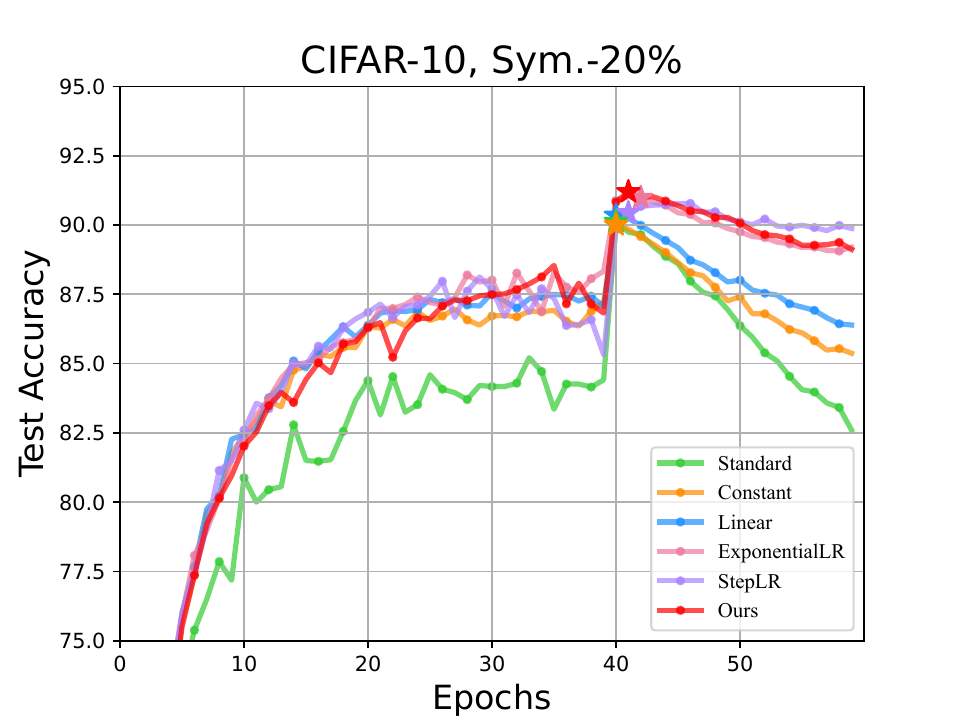}}
    \subfloat[]{
		\label{fig:subfig:cifar10-symmetric-40} 
		\includegraphics[width=0.33\linewidth, height=4.5cm]{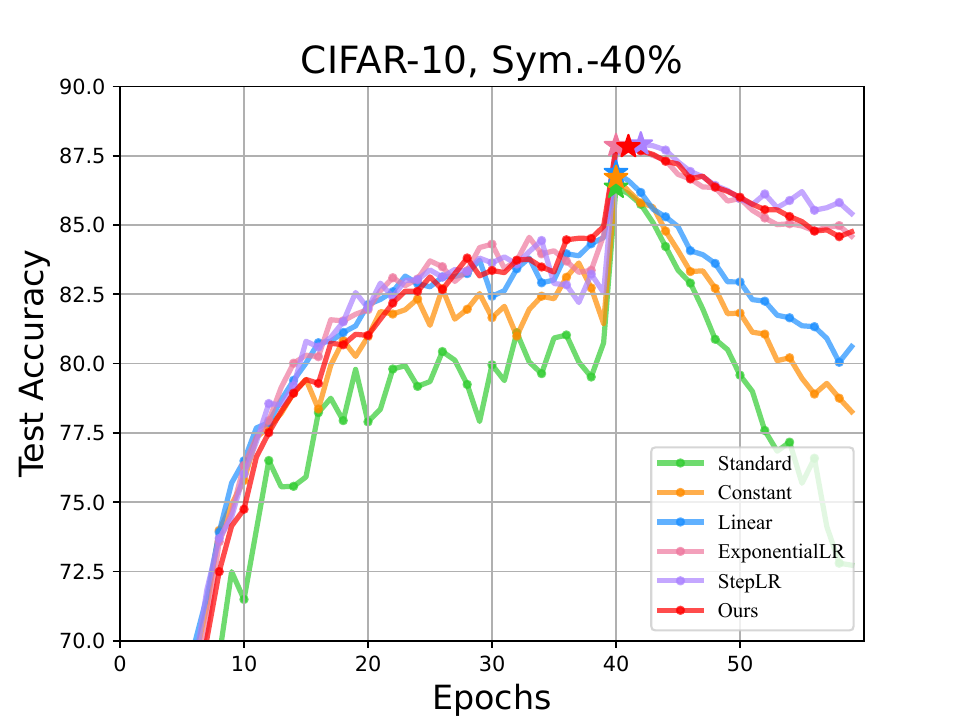}}
    \vspace{3pt}
	\caption{Illustrations of the effect of different $\beta_2$ functions on F-MNIST and CIFAR-10.}
	\label{fig:choose_beta2}
\end{figure*}

\noindent\textbf{Comparison methods.} We compare the proposed method with the following state-of-the-art methods: (1) Standard, which trains deep networks with the cross-entropy loss on noisy datasets. (2) CoTeaching~\cite{han2018co}, which maintains two deep networks and cross-trains on the examples with small losses. (3) CoTeaching+~\cite{yu2019does}, which uses two deep networks and exploits small-loss examples among the prediction disagreement data for training. (4) JoCor~\cite{wei2020combating}, which employs two deep networks and reduces the diversity of the networks to improve robustness and tackle noisy labels. (5) SIGUA~\cite{han2020sigua}, which performs stochastic integrated gradient underweighted ascent to handle noisy labels. We use self-teaching SIGUA in this paper. (6) CNLCU~\cite{xia2022sample}, which incorporates the uncertainty of losses by adopting interval estimation instead of point estimation of losses, to select clean examples. (7) AdaCorr~\cite{zheng2020error}, which corrects incorrect labels by checking the class-posterior ratio between predicted labels and given labels. (8) CDR~\cite{xia2021robust}, which groups all parameters into important ones and redundant ones based on the fitting of clean data. The updates of redundant ones are constrained. Since the proposed method \textbf{t}ackles \textbf{n}oisy \textbf{l}abels with network \textbf{p}arameter \textbf{a}dditive \textbf{d}ecomposition, we name it \textbf{TNLPAD}.
All the implementations are based on PyTorch. Except for the experiments on Clothing1M dataset which are conducted on NVIDIA A100 GPUs, other experiments and conducted on NVIDIA TITAN V GPUs. Note that we do not directly compare our method with some state-of-the-art methods like Self~\cite{nguyen2020self} and DivideMix~\cite{li2020dividemix}, because their proposed methods are aggregations of multiple techniques. The comparison is not fair. Instead, we show that our method can be used to improve cutting edge performance.

\subsection{Comparison with Prior State-of-the-Arts}
\textbf{Results on simulated noisy datasets.} We analyze the experimental results in Table~\ref{tab:simulated_experiments}. Specifically, for MNIST, our proposed method produces the best results in the vast majority of cases. However, our method faces stiffer competition in scenarios with instance-dependent noise, where CoTeaching+ edges ahead. The advantage of CoTeaching+ in this context is attributed to the less complex nature of MNIST, which allows for the effective employment of small-loss trick to accurately sieve out clean samples for training. When the analysis extends to more complex datasets such as F-MNIST, CIFAR-10, and CIFAR-100, the limitations of CoTeaching+ become apparent. It cannot achieve accurate selection and thus does not achieve the best performance.
For F-MNIST and CIFAR-10, our method is superior to other state-of-the-art methods. Although some baselines such as CDR, can achieve good results under some noise settings, it does not handle instance-dependent noise well. In contrast, our method shows good robustness across a wide range of noise settings. 
Finally, for CIFAR-100, our proposed method still performs best and achieves a large performance improvement compared to other state-of-the-art methods, highlighting its robust adaptability and superior handling of various complex noise.
The results verify the effectiveness of our method against simulated label noise.

\noindent\textbf{Results on real-world noisy datasets.} Table~\ref{tab:real_world} shows the experimental results on Food-101, Clothing1M and CIFAR-10N datasets. For Food-101, the proposed method consistently outperforms all baseline methods, which achieves an improvement of 1.41\% over the most advanced baseline method AdaCorr. For Clothing1M, our method also achieves a clear lead over baselines. More specifically, our result provides a 1.89\% improvement over the best baseline method JoCor. Similarly, for CIFAR-10N, our method still performs well in classification accuracy and achieves optimal results. For CIFAR-10N-W, we achieve an improvement of 1.65\% over the optimal baseline method AdaCorr. The results fully demonstrate the effectiveness of our method for real-world label noise.



\subsection{Combining Semi-supervised Learning Methods}
Recently, it is popular to combine technologies in semi-supervised learning to handle noisy labels, \eg, Self~\cite{nguyen2020self}, DivideMix~\cite{li2020dividemix}. We demonstrate that the proposed method can be used to improve cutting edge performance of these state-of-art methods by combining semi-supervised learning methods. Specifically, we combine the proposed method with semi-supervised methods Self and DivideMix, respectively. The experiments are conducted on simulated noisy CIFAR-10. We show the results in Table~\ref{tab:combination_with_semi}. As we can see, the combination with our method brings a certain improvement in classification performance for Self and DivideMix in most cases.

\subsection{More Analyses}

\textbf{Experiments with higher noise levels.} We set the noise rate to 20\% and 40\% before. Here, to verify the robustness under higher noise levels, we set the noise rate to 60\%, 70\%, and 80\%. We exploit symmetric noise because the condition that clean labels are diagonally dominant will be destroyed if the noise rate of the other three kinds of noise is more than 50\%~\cite{ma2020normalized}. Experimental results are provided in Table~\ref{tab:high_noise_levels}. As can be seen, for CIFAR-10, our method outperforms all state-of-the-art methods under various high noise levels. Similarly, for the more challenging dataset, \ie, CIFAR-100, our method achieves varying degrees of lead over the baselines. Especially when the noise rate reaches 80\%, many baseline methods do not work well, but our method can still achieve the highest classification accuracy. This proves the robustness and superiority of our method under high noise levels.

\begin{table}[!t]
    \centering

    \vspace{4pt}
    \renewcommand{\arraystretch}{1.25}
    \caption{Ablation study results in terms of test accuracy (\%) on simulated noisy CIFAR-10 and CIFAR-100. Best experimental results are \textbf{boldfaced}.}
    \begin{tabular}{l|ccc|c|c}
    \hline
    & PD & constrain $\bm{\gamma}$ & constrain $\bm{\sigma}$ & Sym.~20\% & Sym.~40\% \\\cline{1-6}
    \multirow{4}{*}{\rotatebox[origin=c]{90}{CIFAR-10}} 
    & & & & 90.19$\pm$0.31 & 86.26$\pm$0.11\\\cline{2-6}
    & \checkmark & & & 90.09$\pm$0.16 & 86.63$\pm$0.36\\\cline{2-6}
    & \checkmark & \checkmark & & 90.87$\pm$0.22 & 87.68$\pm$0.27\\\cline{2-6}
    & \checkmark & \checkmark & \checkmark & \textbf{91.07$\pm$0.19} & \textbf{87.83$\pm$0.31}  \\
    
    \hline
    \hline
    \multirow{4}{*}{\rotatebox[origin=c]{90}{CIFAR-100}} 
    & & & & 65.77$\pm$0.61 & 60.58$\pm$0.34\\\cline{2-6}
    & \checkmark & & & 66.31$\pm$1.02 & 60.29$\pm$0.66\\\cline{2-6}
    & \checkmark & \checkmark & & 68.39$\pm$0.14 & 63.00$\pm$0.49\\\cline{2-6}
    & \checkmark & \checkmark & \checkmark & \textbf{69.46$\pm$0.25} & \textbf{63.63$\pm$0.38} \\
    \hline
    \end{tabular}
    \vspace{3pt}
    \label{tab:ablation}
\end{table}

\noindent\textbf{Consistent leads with different networks.} To show that our method is stable to the chosen network architectures, we try other popularly used networks, \ie, Wide-ResNet-28~\cite{zagoruyko2016wide}, VGG-19~\cite{simonyan2014very}, and ResNeXt-29~\cite{2016Aggregated}. Except for the network structure, other optimization details remain the same. Experimental results are shown in Table~\ref{tab:networks}. It can be seen that our method outperforms the other baseline methods in almost all scenarios. Note that for the VGG-19 network, the CDR performs better on CIFAR-100 with 40\% symmetric noise. 
However, for the other network structures, \ie, Wide-ResNet-28 and ResNeXt-29, our method obtains the best results.
We perform an in-depth analysis of how the proposed method consistently outperforms other methods under different network structures. Specifically, Our method introduces functional segregation at the parameter level. In LNL, during the network's training phase, incorrect labels can propagate errors through the network during backpropagation, leading to adjustment bias of all trainable parameters. Our method directly addresses this challenge by ensuring that the impact of incorrect labels is confined to the noise-specific parameters, thus safeguarding the parameters responsible for learning from clean data. 

Note that this decomposition is model-independent, meaning it does not depend on the specific architecture of the deep network in use. Therefore, regardless of the network configuration, our method consistently outperforms others because it intuitively reduces the interference of noise on the parameters and enables the network to focus on accurate data representation. This parameter-level control is the underlying reason for the consistent enhancement in performance across different network structures observed with our proposed method.

\noindent\textbf{Ablation study.} As described in the Section~\ref{sec:method}, our method can be generally split into three modules: parameter decomposition~(PD), the constraint on $\bm{\gamma}$, and the constraint on $\bm{\sigma}$. We here conduct ablation studies to explain what makes our method successful. Experiments are performed on simulated noisy datasets CIFAR-10 and CIFAR-100 with symmetric noise. Results are shown in Table \ref{tab:ablation}. It should be noted that when we do not constrain $\bm{\sigma}$ or $\bm{\gamma}$, we use $\mathbf{w}$ for prediction. In other experiments, we make predictions with $\bm{\sigma}$.

First, without parameter decomposition, what we obtain is the result of Standard. 
Then, if we only perform parameter decomposition and do not constrain $\bm{\sigma}$ and $\bm{\gamma}$, it can be seen that the results on the test set are very close to Standard. In other words, parameter decomposition itself does not bring any improvement to network performance. 
Next, if we add the constraint on $\bm{\gamma}$ based on parameter decomposition, the test accuracy is significantly improved compared with Standard. This indicates that by adding the constraint on $\bm{\gamma}$, $\bm{\gamma}$ can absorb the side effect of mislabeled data, which enables $\bm{\sigma}$ to learn as much clean data as possible. 
Furthermore, we add a constraint on $\bm{\sigma}$ based on the constraint on $\bm{\gamma}$. The results show that compared with only constraining $\bm{\gamma}$, adding the constraint on $\bm{\sigma}$ further improves the test accuracy of the network to a certain extent. This demonstrates that constraining $\bm{\sigma}$ allows them to memorize less mislabeled data, and thus obtain better performance.
Overall, ablation studies show that combining the three modules of our method can achieve a steady performance improvement compared with Standard, which proves the necessity of each module in our method and the validity of our proposed TNLPAD.
\begin{figure*}[h]
	\centering

	\subfloat[]{
		\label{fig:subfig:mnist-symmetric} 
        \includegraphics[width=0.32\linewidth, height=4.3cm]{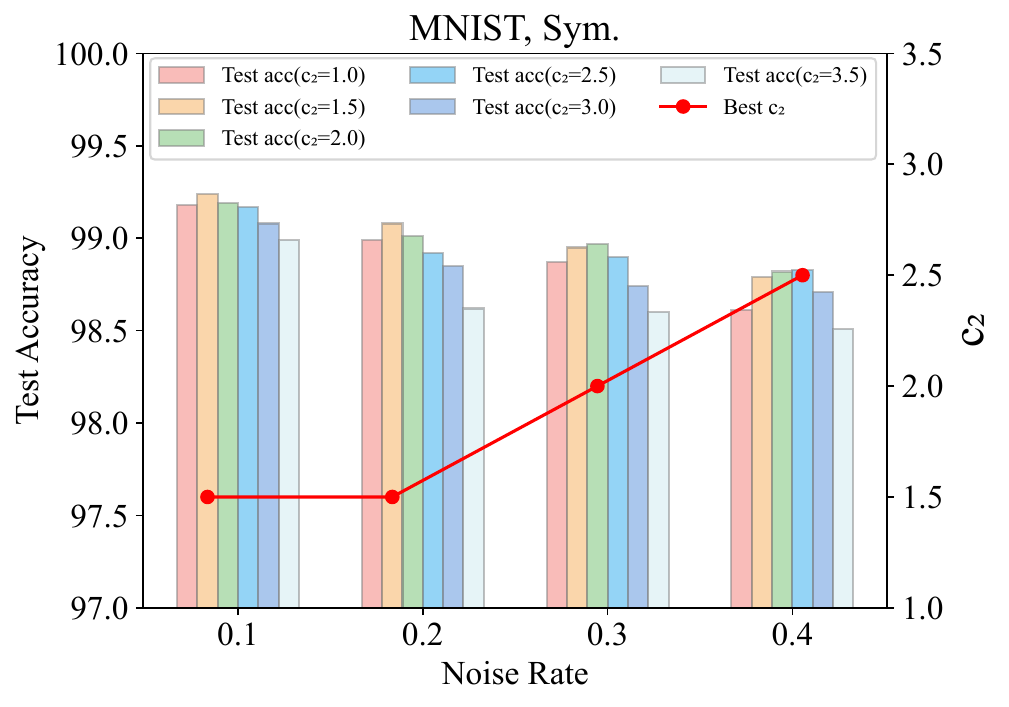}}
    \subfloat[]{
		\label{fig:subfig:F-MNIST-symmetric} 
		\includegraphics[width=0.32\linewidth, height=4.3cm]{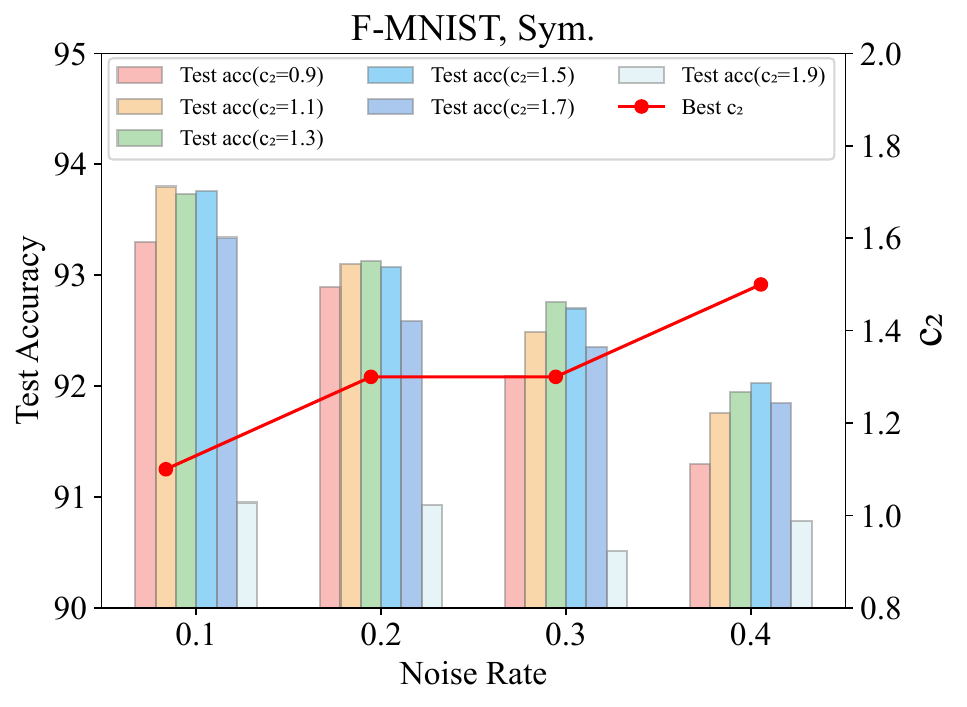}}
    \subfloat[]{
		\label{fig:subfig:cifar10-symmetric} 
		\includegraphics[width=0.32\linewidth, height=4.3cm]{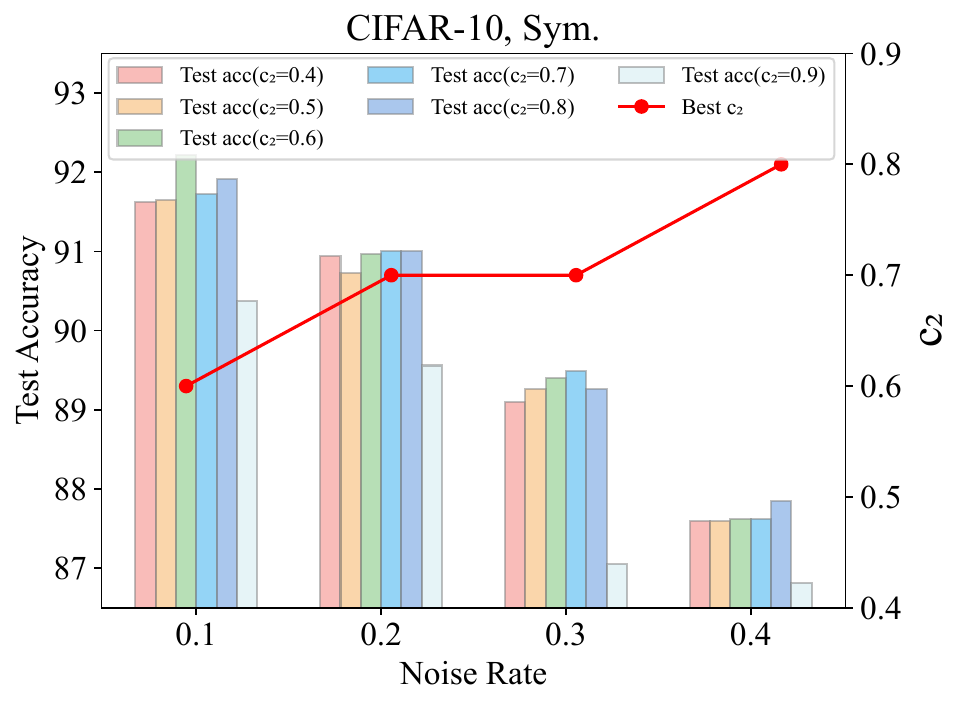}}

    \subfloat[]{
		\label{fig:subfig:mnist-asymmetric} 
		\includegraphics[width=0.32\linewidth, height=4.3cm]{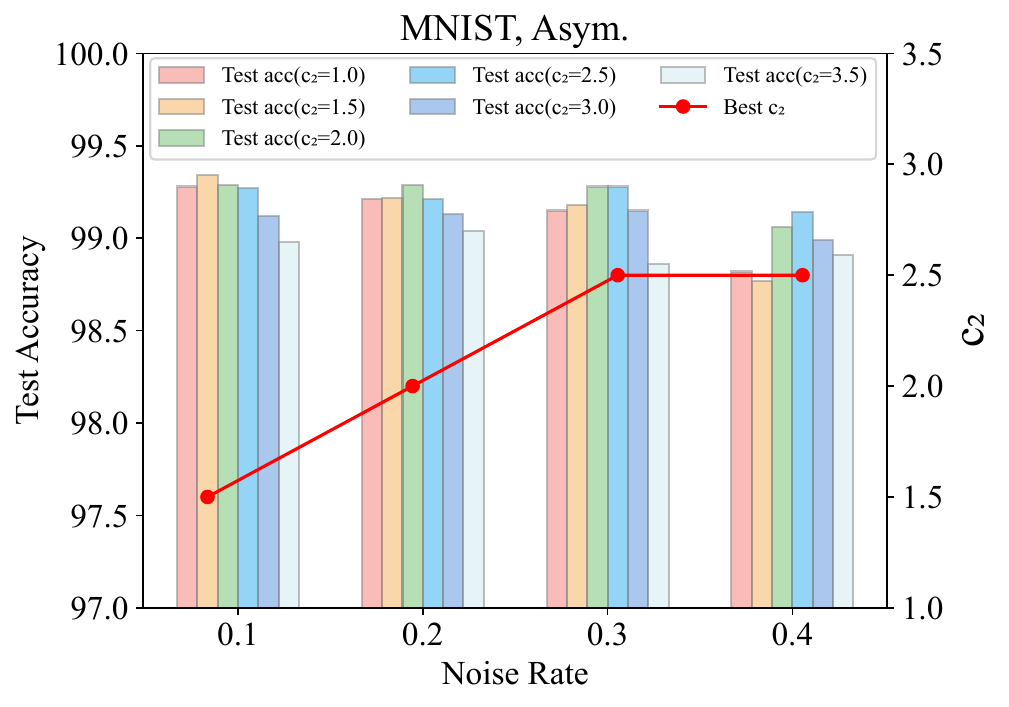}}
    \subfloat[]{
		\label{fig:subfig:F-MNIST-asymmetric} 
		\includegraphics[width=0.32\linewidth, height=4.3cm]{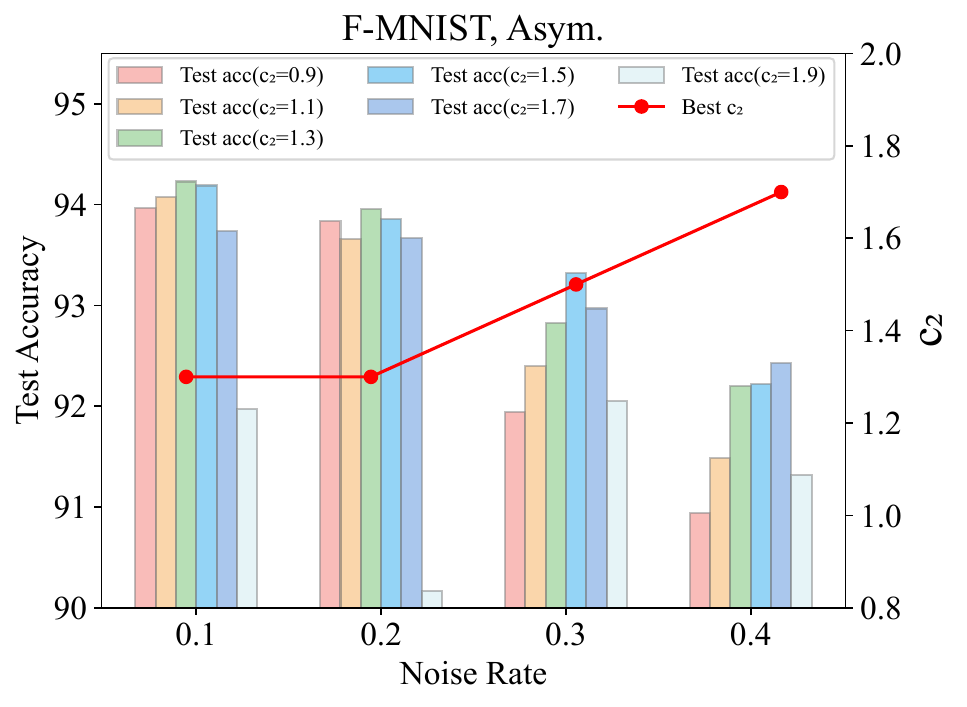}}
    \subfloat[]{
		\label{fig:subfig:cifar10-asymmetric} 
		\includegraphics[width=0.32\linewidth, height=4.3cm]{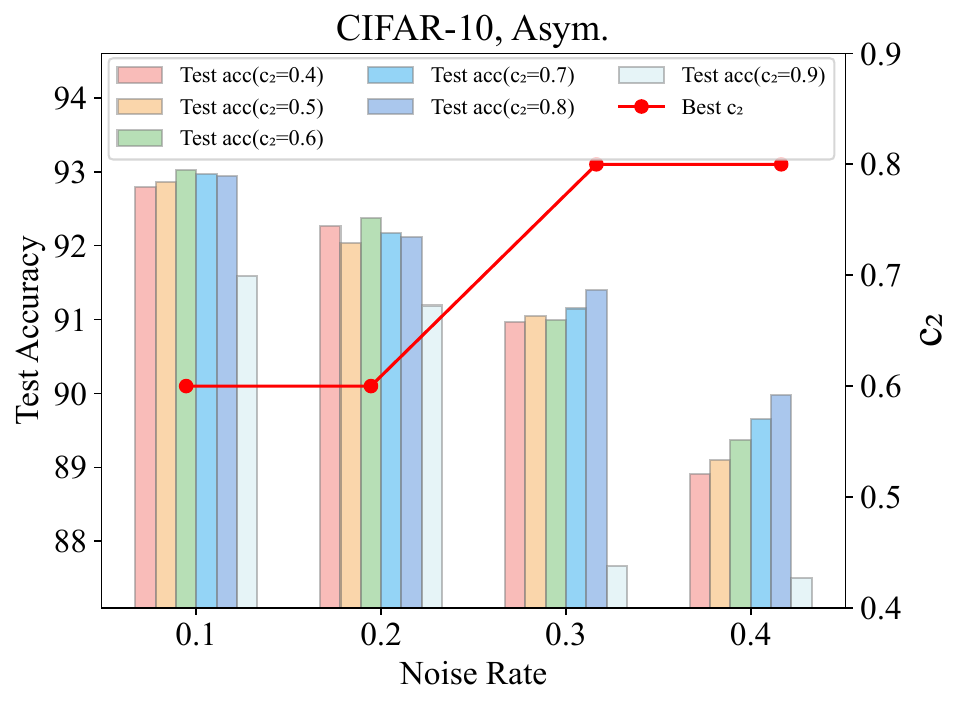}}
    \vspace{3pt}
	\caption{Illustrations of relationship between the hyper-parameter $c_2$ and the noise rate.}
	\label{fig:relationship-c2-noise_rate}

\end{figure*}


\noindent\textbf{The choice of $\beta_2(t)$.} As we discuss in Section~\ref{sec:method}, $\beta_1(t)$ associated with $\bm{\sigma}$ should be an increasing function, while the $\beta_2(t)$ associated with $\bm{\gamma}$ should be a decreasing function. From the results in Table~\ref{tab:ablation}, it can be seen that the constraint on $\bm{\gamma}$ plays an important role in our method. It greatly impacts the performance of TNLPAD. Therefore, in this part, we discuss the impact of different function types of $\beta_2(t)$ on performance. 

We choose several function types including constant function, linear function, power function(ours), ExponentialLR function, and StepLR function, where the ExponentialLR function controls the change of the function value in the form of an exponential function, and the StepLR function adjusts the function value in an equal interval of decreasing. Both of them are implemented with the help of the PyTorch learning rate adjustment strategy. The different function curves varying with epoch $t$ are shown in Fig.~\ref{fig:subfig:decre_function}. We conduct experiments on simulated noisy datasets F-MNIST and CIFAR-10 with symmetric noise of two ratios for full verification. The results are shown in Fig.~\ref{fig:subfig:fmnist-symmetric-20}, Fig.~\ref{fig:subfig:fmnist-symmetric-40}, Fig.~\ref{fig:subfig:cifar10-symmetric-20}, and Fig.~\ref{fig:subfig:cifar10-symmetric-40}.
The star on each curve indicates the network’s best performance on the validation set during training. The Standard is the vanilla early stopping method and we set it as a baseline.

First, we can observe that the result of the constant function is similar to Standard. This is because the constant constraint causes the constraint on $\bm{\gamma}$ to be too large in the later training stage, which prevents them from effectively absorbing the impact of mislabeled data. This demonstrates our idea that the constraint applied on $\bm{\gamma}$ should be a decreasing function.

Second, we can see that the linear function can have better performance than the constant function because it reduces the constraint on $\bm{\gamma}$ during training. However, due to its slow descent of constraints on $\bm{\gamma}$, it cannot perform better than other decreasing functions in most scenarios.

Third, the ExponentialLR function and the StepLR function usually have similar results. Actually, from Fig.~\ref{fig:subfig:decre_function} we can see that the ExponentialLR and StepLR have similar decreasing trends. This indicates that the key to selecting a constraint function in our method is the function’s trend rather than its family. These two functions outperform the linear function in most scenarios because they decay the constraints in a timely manner. However, they still cannot outperform the power function in most scenarios, because their decay rates are too high, resulting in $\bm{\gamma}$ sharing the knowledge with $\bm{\sigma}$ on clean data. 

Finally, the power function can outperform all other functions in most cases. We think this is due to the fact that the changing trend of the power function can well match the training process of the network. From the experimental results, we can also see that all decreasing functions can outperform the Standard. Moreover, in some scenarios, the results of the power function are not far from those of other functions such as ExponentialLR. Therefore, our method is robust to the function family. By carefully adjusting hyper-parameters, many function families can achieve good performance. In this paper, we simply adopt the power function to constrain $\bm{\gamma}$ (\eg, $\beta_2(t)=t^{-c_2}$).
Also, the choice of the hyper-parameter $c_2$ is discussed below.

\begin{figure*}[!h]
	\centering
    \vspace{5pt}
    \subfloat[Pair.-40\%]{
		\label{fig:subfig:tSNE-pairflip} 
        \includegraphics[width=0.48\textwidth]{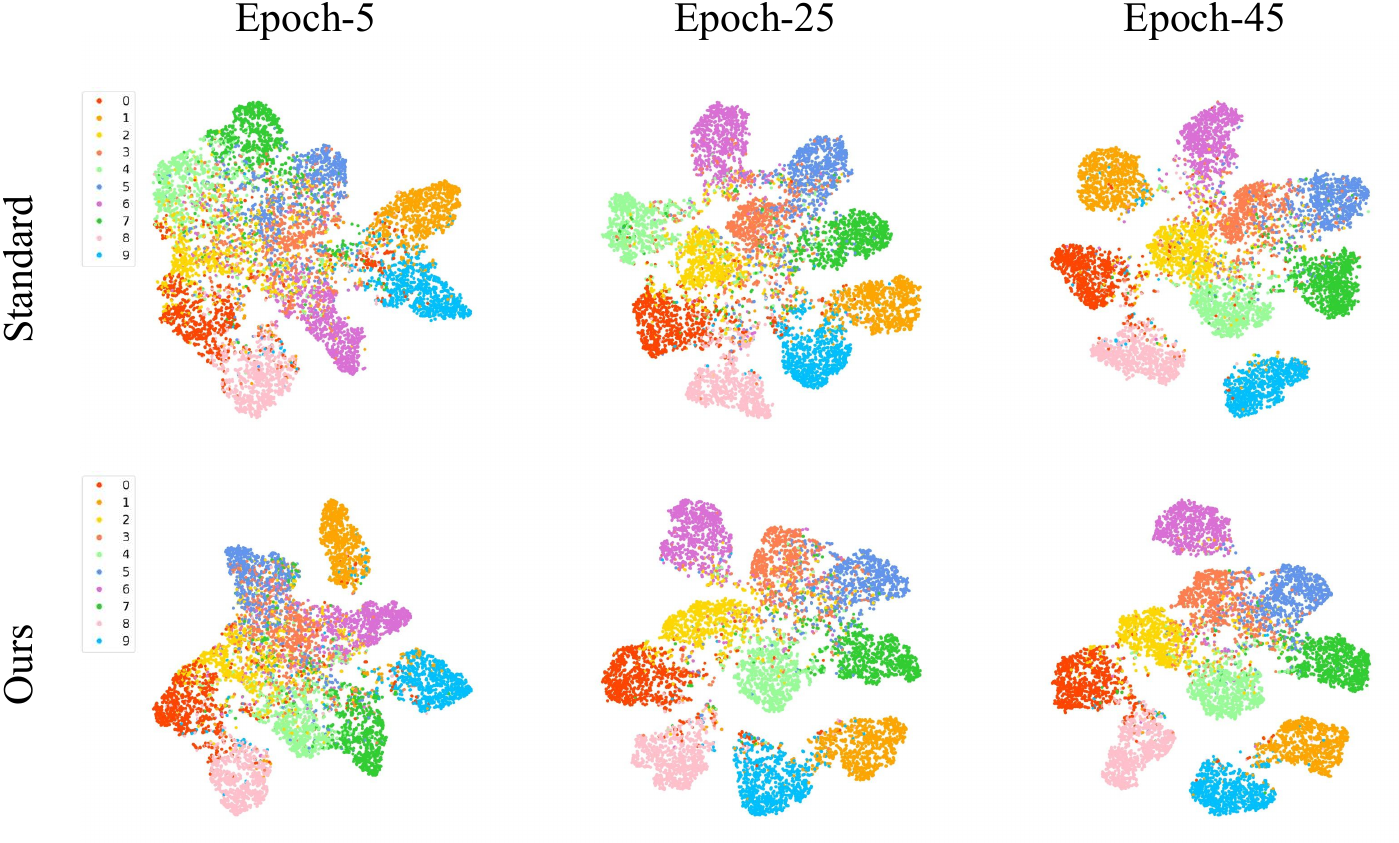}}
    \subfloat[Ins.-40\%]{
		\label{fig:subfig:tSNE-instance} 
		\includegraphics[width=0.48\textwidth]{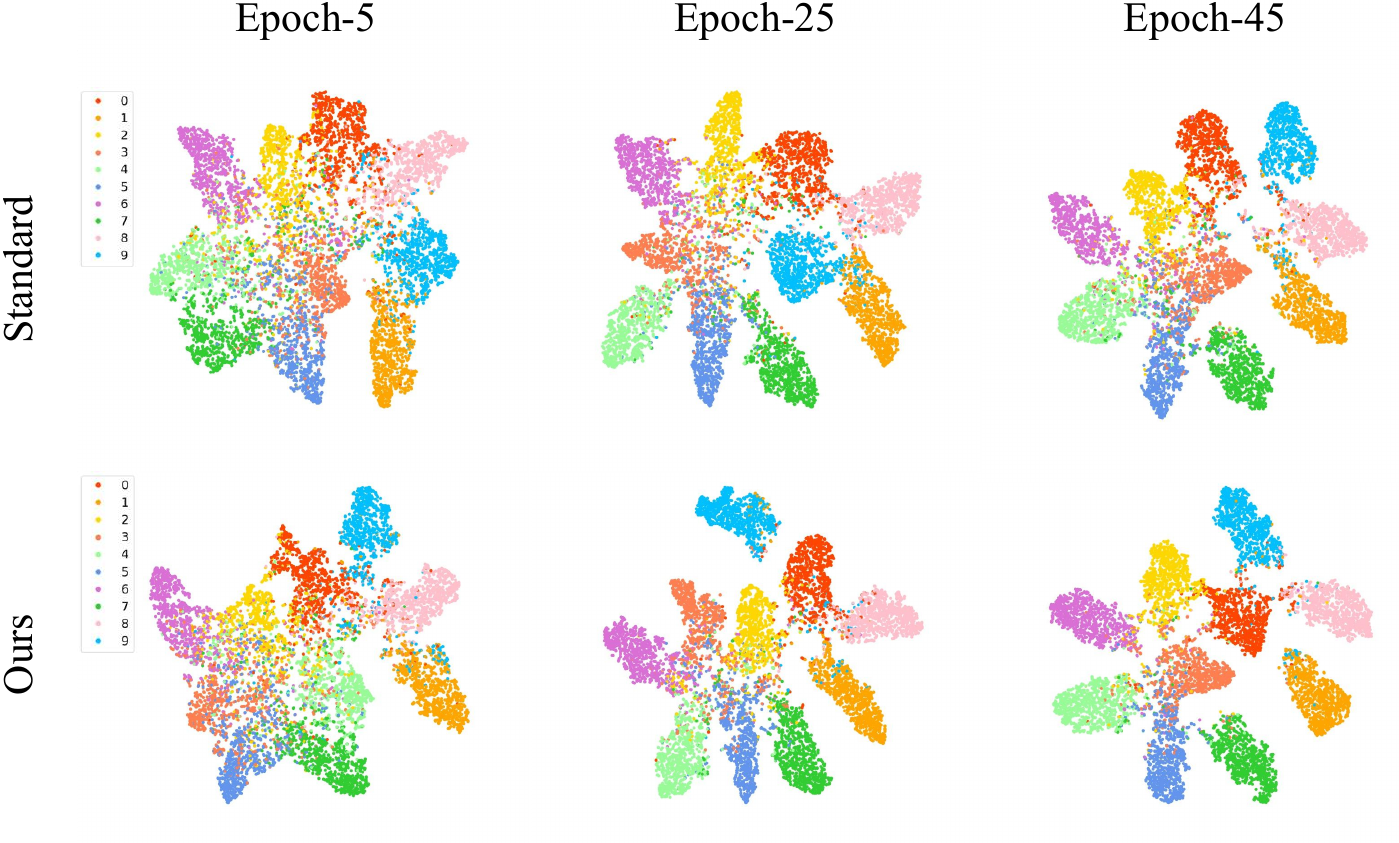}}
    \vspace{3pt}
	\caption{Visualization of experimental results using t-SNE. The experiments are conducted on CIFAR-10.}

	\label{fig:tSNE}
\end{figure*}

\noindent\textbf{Relationship between hyper-parameter and noise rate.} Note that we exploit $\bm{\gamma}$ to absorb the influence of mislabeled data. Therefore, intuitively, the values of the hyper-parameter $c_2$ should be related to the noise rate. We verify this claim empirically. As shown in Fig.~\ref{fig:relationship-c2-noise_rate}, 
we can see that the optimal value of $c_2$ for each noise rate has a positive correlation with the noise rate (shown as the red line). In other words, when the noise rate increases, a larger $c_2$ can make the updates of $\bm{\gamma}$ faster and force $\bm{\gamma}$ to absorb the impact of mislabeled data more effectively. Better network robustness can be hence achieved. It is noteworthy that within a certain range for a given dataset, the selection of $c_2$ does not significantly affect accuracy. Consequently, although there are different optimal $c_2$ settings for each noise rate, we choose only one value of $c_2$ per dataset for simplicity. Specifically, for MNIST, we set $c_2=2.0$. For F-MNIST, we set $c_2=1.5$. For CIFAR-10, we set $c_2=0.6$. For CIFAR-100, we set $c_2=0.2$. Lastly, for Food-101, Clothing1M, and CIFAR-10N, we set $c_2=0.8$.

\noindent\textbf{t-SNE visualization on representations.} We visualize the experimental results using t-SNE~\cite{van2008visualizing}, which is a technique developed by van der Maaten and Hinton in 2008. To initiate this process, high-dimensional features are extracted from the second-to-last layer of the network, chosen for its rich representational capacity just before the final classification layer. These features encapsulate the learned abstractions of the input data, which are crucial for understanding the model's internal representations.
Following the extraction, t-SNE is employed to translate the high-dimensional feature space into a two-dimensional plane. This dimensionality reduction is performed on the entirety of the test dataset's features, facilitating the visualization of data distribution and cluster formation. In the resulting 2D t-SNE maps, each point corresponds to a test image, with the proximity between points reflecting the similarity of their high-dimensional features.

Fig.~\ref{fig:tSNE} shows the results of the visualization on CIFAR-10 with two types of noise at different training epochs. Specifically, At epoch 5, Standard and the proposed method demonstrate a blend of class features, indicating early-stage training where class separation is not yet distinct. As training progresses to epoch 25 and further to epoch 45, the proposed method exhibits a marked improvement in class distinction compared with Standard. The clusters become more defined and segregated, highlighting the ability of the proposed method to distinguish between different classes in noisy data. The plots clearly illustrate that the proposed method achieves a more refined separation, pointing to robustness against noisy labels and an enhanced ability to extract features, which is crucial for accurate classification in practical scenarios with imperfect data.

\section{Conclusion}\label{sec:conclusion} 

In this paper, we focus on tackling noisy labels from the perspective of network parameters. A novel method based on additive parameter decomposition is proposed, where all parameters are decomposed into the parameters for memorization of clean data and memorization of mislabeled data respectively. We restrict the former to fit clean data, while the latter to absorb the side effect of mislabeled data. The robustness of deep networks can be improved by only utilizing the parameters for memorization of clean data. Extensive experiments demonstrate the effectiveness of the proposed method. 
We are interested in deriving theoretical analysis for our method. For future work, we are particularly interested in assessing the viability and effectiveness of our method in other complex domains, such as natural language processing. This expansion will allow us to understand the broader implications of our method and refine it for diverse challenges in machine learning.

\section*{Acknowledgments}
This work was supported by the National Natural Science Foundation of China (No. 62376282, No. 62372459).

{ 
\tiny
\bibliography{bib}
}

\end{document}